\documentclass[a4paper,fleqn]{cas-sc}

\usepackage[authoryear,longnamesfirst]{natbib}
\usepackage{subfig}
\usepackage{tabularx}
\usepackage{hyperref}
\usepackage{xurl}
\newcolumntype{L}{>{\raggedright\arraybackslash}X} % column type used in tab:explanations
\usepackage[authoryear]{natbib}

\def\tsc#1{\csdef{#1}{\textsc{\lowercase{#1}}\xspace}}
\tsc{WGM}
\tsc{QE}
\tsc{EP}
\tsc{PMS}
\tsc{BEC}
\tsc{DE}
\begin{document}
\let\WriteBookmarks\relax
% Float placement parameters (the template's \floatpagepagefraction and
% \textpagefraction names are misspelt and have no effect)
\renewcommand{\topfraction}{0.9}
\renewcommand{\bottomfraction}{0.8}
\renewcommand{\textfraction}{0.07}
\renewcommand{\floatpagefraction}{0.7}
\setcounter{topnumber}{3}
\setcounter{bottomnumber}{2}
\setcounter{totalnumber}{4}
\shorttitle{Supervised End-to-End Driving on Public Roads}
\shortauthors{J.S. Berrio-Perez et~al.}
%\begin{frontmatter}

\title [mode = title]{Supervised End-to-End Driving on Public Roads: \\ Data, Reproducible Validation, and Human Supervision}                      
%\tnotemark[1,2]

%\tnotetext[1]{This document is the results of the research
%   project funded by the National Science Foundation.}

%\tnotetext[2]{The second title footnote which is a longer text matter
 %  to fill through the whole text width and overflow into
 %  another line in the footnotes area of the first page.}

\author[1]{Julie Stephany Berrio-Perez}[orcid=0000-0003-3126-7042]
\cormark[1]
\ead{s.berrioperez@uq.edu.au}

\author[2]{Eduardo Nebot}[orcid=0000-0002-3914-3741]
\ead{eduardo.nebot@sydney.edu.au}

\author[3]{Javier Ibanez-Guzman}[orcid=0000-0002-5300-9709]
\ead{javier.ibanez-guzman@ampere.cars}

\author[4]{Nick Reed}[orcid=0000-0002-7794-6123]
\ead{nick@reed-mobility.co.uk}

\author[5]{Miguel Angel Sotelo}[orcid=0000-0001-8809-2103]
\ead{miguel.sotelo@uah.es}

\author[6]{Cristina Olaverri-Monreal}[orcid=0000-0002-5211-3598]
\ead{cristina.olaverri-monreal@jku.at}

\author[1]{Zuduo Zheng}[orcid=0000-0002-5289-2106]
\ead{zuduo.zheng@uq.edu.au}

\author[7]{Prathyush P Menon}[orcid=0000-0003-3804-9291]
\ead{P.M.Prathyush@exeter.ac.uk}

\author[8]{Walter Zimmer}[orcid=0000-0003-4565-1272]
\ead{wz@ucla.edu}

\author[9]{Juan Medina-Lee}[orcid=0000-0003-4489-4280]
\ead{juan.medina26@upr.edu}

\author[10]{Mona Mirzaie}[orcid=0009-0002-6501-5038]
\ead{mirzaie@tnt.uni-hannover.de}

\author[1]{Tyson Phillips}[orcid=0000-0001-8729-0693]
\ead{t.phillips1@uq.edu.au}

\author[2]{Mao Shan}[orcid=0000-0002-5032-6581]
\ead{mao.shan@sydney.edu.au}

\affiliation[1]{organization={The University of Queensland},
                city={Brisbane},
                country={Australia}}

\affiliation[2]{organization={The University of Sydney},
                city={Sydney},
                country={Australia}}

\affiliation[3]{organization={Renault Group},
                country={France}}

\affiliation[4]{organization={Reed Mobility},
                country={United Kingdom}}

\affiliation[5]{organization={University of Alcal\'a},
                city={Alcal\'a de Henares},
                country={Spain}}

\affiliation[6]{organization={Johannes Kepler University Linz},
                city={Linz},
                country={Austria}}

\affiliation[7]{organization={University of Exeter},
                city={Exeter},
                country={United Kingdom}}

\affiliation[8]{organization={University of California Los Angeles},
                city={Los Angeles},
                country={USA}}

\affiliation[9]{organization={University of Puerto Rico--Mayag\"uez},
                city={Mayag\"uez},
                country={Puerto Rico}}

\affiliation[10]{organization={Leibniz University Hannover},
                 city={Hannover},
                 country={Germany}}

\cortext[1]{Corresponding author}

%\nonumnote{This note has no numbers. In this work we demonstrate $a_b$
  % the formation Y\_1 of a new type of polariton on the interface
  % between a cuprous oxide slab and a polystyrene micro-sphere placed
  % on the slab.
  % }

\begin{abstract}
Supervised end to end (E2E) driving systems can now perform most of the dynamic driving task in production vehicles on public roads, in both left hand and right hand traffic. Although these systems may control the vehicle for most of a trip, the human driver remains legally responsible and must supervise their operation. This perspective paper argues that this transition changes the evidence needed to demonstrate safety. Much of the relevant evidence, including fleet driving logs, intervention records, driver state data and the behaviour of learned policies, is proprietary. Consequently, safety claims are difficult to verify independently or reproduce.
We examine three related challenges. First, the transition from modular pipelines to learned policies changes how systems must be verified and validated. We illustrate this challenge using intelligent speed assistance and the objective evaluation of lane support systems during normal operation. Second, good driving must be defined using measures that go beyond fatality statistics, which cannot establish whether automated driving is consistently careful and competent. Safety assurance should therefore be underpinned by standardised in-vehicle behavioural data. Third, concerns about driver engagement, excessive trust and interface design mean that the human and AI system, rather than the model alone, must be treated as the unit of safety. 
We map modular architectures, policies developed through imitation learning and policies refined through reinforcement learning to the data required for their evaluation. We also consider prediction and causal reasoning modules, explanation methods, driver monitoring and behavioural assurance logging. We then review the open datasets, tools and benchmarks available to researchers and identify the remaining gaps. We conclude that hybrid architectures combining learned policies with independently inspectable components, supported by open data, open source tools and reproducible evaluation practices, offer the most credible path towards transparent assurance. We distil these findings into a research agenda.
\end{abstract}

%\includegraphics{figs/cas-grabs.pdf}

 % \begin{highlights}
 % \item Supervised end-to-end driving shifts safety evidence from physics to data
 % \item Fatality statistics cannot show careful and competent automated driving
 % \item The human--AI pair, not the model, is the unit of safety
 % \item We map driving architectures to open data, tools and benchmarks
 % \item We propose reporting practices for reproducible safety claims
 % \end{highlights}

\begin{keywords}
Autonomous vehicle navigation, end-to-end driving, intelligent transportation systems, human factors and human-in-the-loop, imitation learning, datasets for robot learning, performance evaluation and benchmarking.

\end{keywords}

\maketitle

\section{Introduction}
\label{sec:introduction}

Lane support, adaptive cruise control, and automatic emergency braking are fitted to a large share of new vehicles, and in 2025 roughly one third of the cars sold worldwide offered some form of SAE Level~2 driving assistance \citep{sae2021j3016, berginsight2026drivingautomation, iea2026autonomousvehicles}. At the same time, the technology underneath these systems is changing in a way that the established engineering and regulatory frameworks were not designed for.
For two decades, automated driving was built as a modular pipeline: sensing, perception, prediction, planning, and control, each stage engineered, tested, and certified largely on its own, with behaviour dictated by written requirements \citep{iso26262,iso21448}. That system design delivered successful deployments in structured domains such as stevedoring and mining, but stalled in urban traffic, where the number and unpredictability of interacting agents made intention prediction, safety demonstration, and sensor cost unmanageable.
 
Since roughly 2024, a different approach has reached series production. End-to-end (E2E) driving models learn a direct mapping from sensor and navigation inputs to driving actions \citep{bojarski2016end,hu2023planning,chen2024end}. Public presentations by industry developers indicate that production systems can combine large-scale imitation learning with refinement using fleet-mined scenarios and reinforcement-learning techniques \citep{lu2022imitation, elluswamy2025building}.

Deployed as \emph{supervised} E2E driving (sometimes marketed as ``L2++'') such systems can perform the majority of the dynamic driving task from origin to destination on public roads in left- and right-hand traffic, while the human in the driver's seat remains legally responsible and is required to keep eyes on the road and intervene when necessary. Unlike conventional lane-keeping functions, which many owners never turn on, these systems are used for the vast majority of trip time, and several manufacturers and suppliers have announced comparable products.
 
The vehicle has become a reasoning agent operating in an open world, and its behaviour is implicitly defined by training data rather than explicitly specified.
Validation can no longer be purely physics-based and deterministic; it is probabilistic, and the number of kilometres required to show safety statistically is far beyond what any test programme can drive \citep{kalra2016driving}. Explainability, over-the-air updates, the certification of simulators and tools, and operational design domains that cannot in practice be enforced all become part of the validation problem \citep{iso8800}. On the human side, the driver's role has shifted from operator to supervisor, a role that decades of human-factors research warn is poorly suited to people: prolonged, reliable automation breeds hypovigilance and over-trust, takeover times are task-dependent and often too long, and users actively circumvent attention monitoring \citep{bainbridge1983ironies,olaverri2020}. Meanwhile, the everyday Advanced Driver Assistance Systems (ADAS) functions that already ship in millions of vehicles are still assessed mainly on closed tracks or through subjective driver ratings, so their real-world effectiveness (and the large differences between implementations, even within one manufacturer's line-up) remain largely unmeasured.
 
%This paper summarises the presentations and discussion of the workshop \emph{Introducing the Next Level of ADAS on the Roads}, held at the IEEE International Conference on Intelligent Transportation Systems (ITSC) 2026 in Naples. The workshop reunited academic, industrial and regulatory perspectives around the statement: what does it take to test, validate and supervise the next level of driver assistance on real roads? Rather than reporting the talks in sequence, we organise their content along themes that emerged as the day's common thread:

This paper is a perspective on what it takes to test, validate and supervise the next level of driver assistance on real roads, and on why answering that question has become a matter of open science.\footnote{This paper was inspired by, and builds upon, the presentations and discussions at the workshop \emph{Introducing the Next Level of ADAS on the Roads}, held at the IEEE International Conference on Intelligent Transportation Systems (ITSC) 2026 in Naples, Italy.} Drawing on academic, industrial and regulatory perspectives, we organise the argument along four themes:

\begin{enumerate}
  \item the transition from modular to end-to-end architectures and what it changes for verification and validation, illustrated with a practitioner's account of validating a deceptively easy function, intelligent speed assistance;
  \item why fatality statistics cannot validate that an automated vehicle is a careful and competent driver, why the public judges safety by behaviour, and how standardised in-vehicle data (Digital Commentary Driving) could make that behaviour assessable;
  \item safety of the combined human–AI system, driver monitoring, and human–machine interface design that keeps drivers appropriately informed without overwhelming them with information or distracting them through excessive notifications.
  \item the data, open-source tools and reproducible verification methods that each architecture requires, what is available to the research community, and what is missing.

\end{enumerate}
 
From these we distil a set of open knowledge gaps (how to define ``good driving'' and against which human baseline safety should be judged, wether the supervisor role is acceptable and how it can be monitored, how E2E behaviour can be explained and assured, and what infrastructure, data-sharing and regulatory changes deployment at scale will require) and propose a research agenda. We argue that the field is moving towards hybrid architectures in which learned driving policies are complemented by explainable prediction, robust driver supervision and continuous, objective in-service evaluation, and that the assurance of such systems depends on data and tools open to independent scrutiny.

The rest of this document is organised as follows. Section~\ref{sec:background} explains the transition from modular pipelines to supervised end-to-end driving. Section~\ref{sec:validation} describes the challenges for validation and verification, including operational-domain definition and in-service evaluation. Section~\ref{sec:gooddriving} considers what constitutes good driving and how behavioural assurance can be supported through interpretable prediction and standardised behavioural evidence. Section~\ref{sec:human} examines the human supervisor and human–machine interface. Section~\ref{sec:openscience} maps the architectures to their data requirements and reviews open data, tools and benchmarks. Section~\ref{sec:discussion} turns the  knowledge gaps into a research agenda, and Section~\ref{sec:conclusion} concludes.

\section{From Modular Pipelines to End-to-End Driving}
\label{sec:background}

%\subsection{The Modular Stack and Where It Stalled}
%\label{sec:background:modular}

The system design that carried automated driving from research to its first deployments is the modular pipeline shown in Fig.~\ref{fig:pipelines} (modular pipeline). A suite of sensors (lidar, radar, cameras, GNSS and inertial units) feeds a perception stage whose job is to build a human-readable representation of the world: a map, the ego-pose within it, and a list of tracked objects with classes, positions, and velocities \citep{11577855}. Downstream, prediction estimates where those objects will be over the next few seconds, planning selects a trajectory, and a controller tracks it. Each stage has its own inputs, outputs, requirements, and test procedures, and the system as a whole can be argued safe by composing arguments about its parts \citep{electronics15143160}. This is the philosophy that functional-safety standards codify, from the generic IEC~61508 \citep{iec61508} to its automotive adaptation ISO~26262 \citep{iso26262}: fixed hardware configurations and behaviour dictated by written requirements.

The approach worked well wherever the environment could be brought under control. Core techniques were developed in the late 1990s and early 2000s in field robotics and produced commercially successful autonomy in stevedoring and mining \citep{durrant2006simultaneous,nebot2008}. Those domains have an operator that owns the workspace.
Autonomous straddle carriers ran in areas from which people were excluded; mining trucks operated in environments that, while more complex, were physically structured for them, with all the humans present trained to co-exist with the machines. Under these conditions, the pipeline could be validated using conventional engineering methods, and the remaining risks could be identified and managed.

The expectation in the early 2010s was that urban driving would follow, and public forecasts placed autonomous cars on city streets by 2019--2020, but this did not happen. Urban traffic contains many more participants whose intentions must be inferred rather than commanded, none of whom can be trained or excluded; the long tail of edge cases made acceptably safe performance difficult to achieve; and the sensor suites needed to make the pipeline robust were too expensive and too demanding in calibration and maintenance to fit into a production vehicle. The modular stack reached the limits of what could be specified in advance.

\subsection{Two Ways Out: Robotaxis and Supervised End-to-End}
\label{sec:background:transition}

Two developments changed this situation within a few years. The first was the maturation of SAE Level~4 robotaxi services, which retained the sensor-rich design but limited operation to a geofenced operational design domain (ODD), with costs that could be supported by a fleet operator \citep{kusano2024comparison}. The second, and the one that concerns this paper was a step change in \emph{supervised} driving assistance brought about by end-to-end learning.

In Austin, Texas, in mid-2025, a robotaxi service built on the rich-sensor pipeline began to operate in the same city as a service built on unmodified production cars equipped with cameras only and an end-to-end policy \citep{flaherty2026technology, brolley2026waymo}. Both had remote-assistance capability; the camera-only vehicles initially carried a safety attendant who was later removed. Although the comparison was not conducted under controlled conditions and does not establish the superiority of either approach, both demonstrated the capability to execute the dynamic driving task effectively and safely in dense urban traffic. An important distinction of the production-vehicle approach, however, is its scalability beyond dedicated robotaxi operations. Since the same software runs on vehicles across the manufacturer's fleet, this capability is also available, in supervised form, in hundreds of thousands of privately owned cars in North America, Australia, New Zealand, and parts of Europe. Several other manufacturers and suppliers have announced comparable systems for 2026–2027. Throughout this paper, we therefore treat Tesla's Full Self-Driving (Supervised) as a case study of a broader category of driving automation, rather than as a singular product.

\subsection{What ``Supervised End-to-End'' Means}
\label{sec:background:definition}

\emph{End-to-end} describes an architecture: a single learned model that maps raw sensor input (typically multi-camera video, plus navigation route and vehicle kinematics) directly to control outputs \citep{bojarski2016end,tampuu2022survey,chen2024end} (Fig.~\ref{fig:pipelines} (End-to-End architecture)). The SAE levels describe the allocation of the dynamic driving task and of the fallback between human and system \citep{sae2021j3016}. An end-to-end policy can be deployed at Level~2, 3, 4 or 5. \emph{Supervised} end-to-end driving is the Level~2 case: the model performs the driving task while a human continuously monitors the environment and is responsible for intervention. Marketing labels such as ``L2+'' or ``L2++'' describe enhanced capabilities but do not constitute additional SAE levels or alter the allocation of responsibility between the human and the system, and it is becoming clear that the absence of agreed definitions between L2, L2++ and L3 is itself a validation problem. The September 2026 revision of SAE J3016 (technically equivalent to ISO TS 22736) states that levels are integers, nominal and mutually exclusive, and must not carry decimals or plus signs. Differences such as hands-off operation are expressed with qualifiers like ‘hands-off capable’ \citep{SAEJ3016_2026}. In J3016 terms a supervised end-to-end system is a Level 2 driver support feature.

From the driver's perspective, supervised end-to-end driving differs from conventional lane-centring and adaptive cruise control in three important ways. First, rather than simply maintaining speed and lane position, the driver selects a destination and the system manages the driving task throughout the journey, including lane changes, intersections, roundabouts, unprotected turns, and navigation through parking areas. Second, steering, acceleration, and braking are coordinated and continuously adapted to road geometry, traffic conditions, and the surrounding environment. For example, the vehicle slows for speed humps and tight curves and accelerates smoothly afterwards, producing driving behaviour that users often describe as natural. Third, these capabilities fundamentally change how drivers interact with automation. While conventional driver-assistance systems are often used only in specific situations, supervised end-to-end systems can be engaged for most of a journey, with drivers intervening mainly when necessary or when taking over specific manoeuvres, such as parking. This shift has important safety implications: the human driver is no longer occasionally assisted by automation but becomes the continuous supervisor of a system performing most of the driving task. Maintaining driver attention, situational awareness, and readiness to intervene therefore becomes a central challenge in the safety of the combined human–AI system.

\subsection{How the Models Are Built}
\label{sec:background:training}

Although the implementation details are proprietary, publicly available information suggests that major developers follow a similar two-phase training approach. Understanding these two phases is important for the discussion of system validation and human factors that follows.

\textit{Phase 1: imitation.}
A base policy is trained by behaviour cloning on large volumes of human driving logged by the fleet, filtered to retain ``good'' driving, and augmented with labelled synthetic data. This yields a model that drives competently in situations well represented in the data, but is brittle in unseen ones and subject to compounding error when small deviations take it outside the training distribution \citep{codevilla2019exploring}. Developers differ mainly in where the data come from: those with a large deployed fleet log it directly, others rely more heavily on simulation.

\textit{Phase 2: refinement on the long tail.}
The base model is then hardened using reinforcement learning and targeted data. The fleet itself is the discovery mechanism: small on-board detector networks flag interventions, disengagements, near-misses and large state changes; these episodes are uploaded, analysed, and used both directly for retraining and as seeds for simulated variations \citep{tesla_fsd_evidence_dashboard}. Because a large fleet can collect huge amounts of driving data every day, it is impossible to retrain on all of it. Instead, the system identifies and uses the most important or unusual cases for training.
Reinforcement learning is also used to induce behaviours that cannot be enforced by imitation, for example widening the margin to pedestrians that human drivers pass too closely.

The models are large by automotive standards and are growing quickly: the case-study developer's first end-to-end release in early 2024 was succeeded within the year by a model with roughly three times the parameter count \citep{electrek_v13_roadmap_2024}, which became the widely used version and the basis of its robotaxi service, and in late 2025 by a further ten-fold increase \citep{notateslaapp_v15_2026}. The perception front end ingests several cameras at tens of frames per second, tokenises the video, and processes a sequence of past context so that the output is conditioned on the recent history rather than a single frame \citep{malik2026teslafsd, teslaFSDsupervisedAustralia}. Such high-frequency processing allows the system to detect and respond rapidly to changing road and traffic conditions in multiple directions, potentially reacting to emerging hazards faster than a human driver.

\subsection{Implications for the Rest of the Paper}

\label{sec:background:consequences}

\begin{figure}[pos=htbp]
  \centering
   \includegraphics[width=0.8\linewidth]{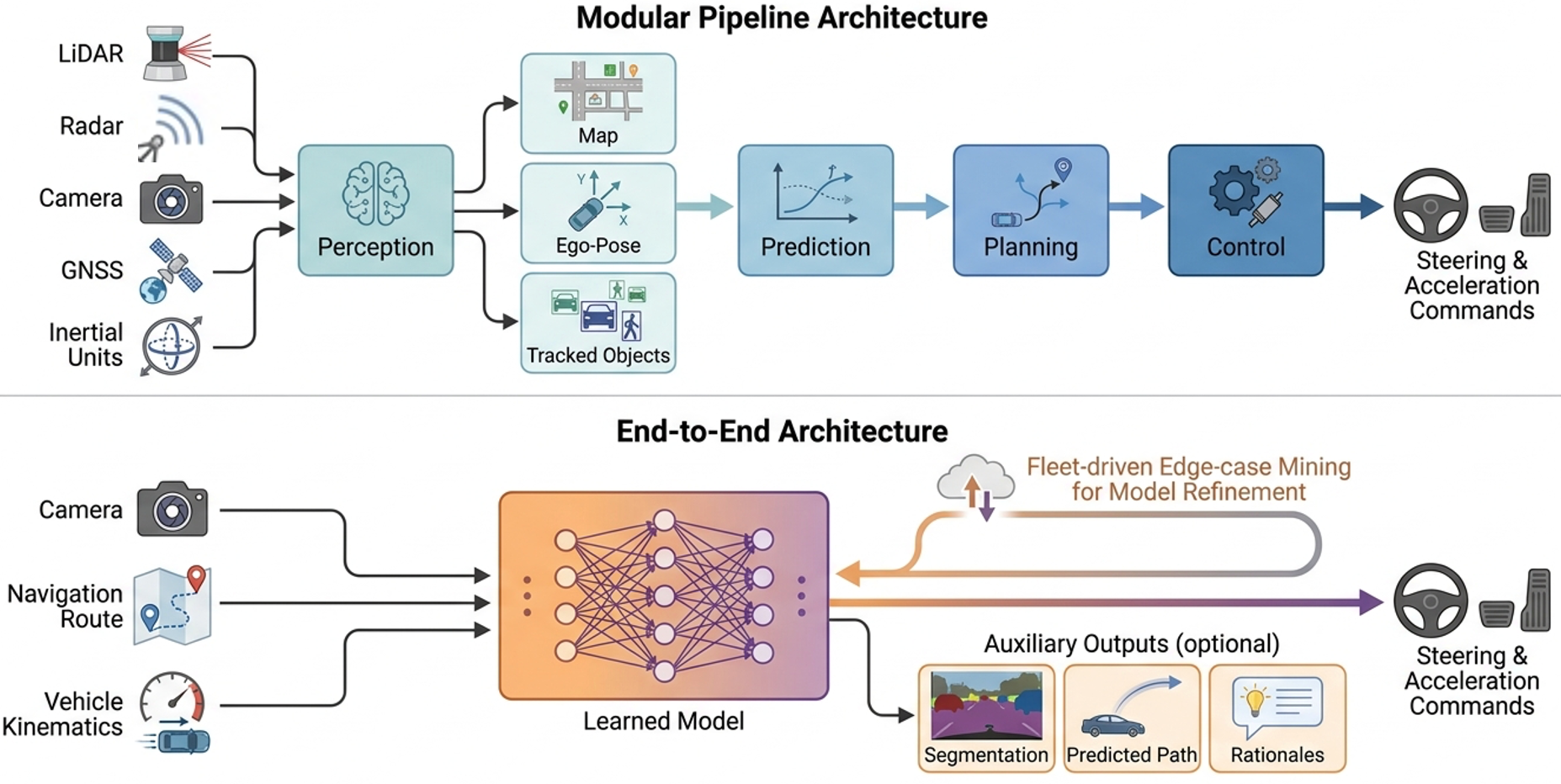}
  \caption{Modular and end-to-end driving architectures.}
  \label{fig:pipelines}
\end{figure}

The change from Fig.~\ref{fig:pipelines} (modular pipeline architecture) to (End-to-End architecture) is a profound change in the scientific foundations of the field. Classical automotive engineering rested on mechanics, structural materials and linear control: deterministic disciplines, with behaviour specified by requirements and validated by physical test. An end-to-end vehicle is a reasoning agent whose behaviour is implicitly defined by its training data, operating in an open world; its natural disciplines are statistical learning, human factors and uncertainty quantification, and validation becomes probabilistic \citep{iso8800,iso21448}. 

The component-level evidence no longer composes into a system argument when there are no components in the traditional sense, and the modern reliance on software-defined vehicles and over-the-air updates means that the object being validated changes continuously (Section~\ref{sec:validation}).

End-to-end systems can and do provide auxiliary outputs (panoptic segmentation, predicted paths, attention or saliency maps, and increasingly natural-language rationales for a manoeuvre) which are essential both for developers analysing failures and for communicating intent to the supervising driver. Whether such rationales are faithful to the policy's actual computation remains and open question, and this is where explainable, causally grounded prediction modules retain a role even inside an end-to-end vehicle (Section~\ref{sec:prediction}).

A system that drives well for hours at a time and expects the human to catch its rare failures places the driver in the supervisory role that human-factors research has warned against since Bainbridge's ``ironies of automation'' \citep{bainbridge1983ironies}. The evidence that end-to-end driving is, in many respects, more attentive than a human (faster reaction, no fatigue, full-surround sensing) makes the discussion of how to keep the supervisor engaged harder, not easier (Section~\ref{sec:human}). Since these advanced systems are deployed in production vehicles alongside conventional driver-assistance features, such as lane-keeping and adaptive cruise control, it is important to objectively evaluate the real-world performance and safety of all these systems, from basic driver assistance to supervised end-to-end driving  (Section~\ref{sec:evaluation}).

\section{Validation and Verification in a Probabilistic World}
\label{sec:validation}

The change of system design described in Section~\ref{sec:background} is  a change in what counts as evidence. For most of the automotive era the object of validation was a mechanical actuator commanded by a human: the driver was the reasoning agent, the vehicle obeyed Newtonian mechanics, thermodynamics and linear control, and safety could be argued deterministically from physics and from requirements. Functional safety formalised in ISO~26262 assumes fixed hardware, decomposable components and behaviour that can be written down before it is built \citep{iso26262}.

\begin{table}[pos=htbp]
  \caption{Evolution of validation metrics and standards for driver assistance and automated driving.}
  \label{tab:metrics}
  \centering
  \small
  \begin{tabular}{@{}p{3.0cm}p{4.8cm}p{4.5cm}@{}}
    \toprule
    \textbf{Generation} & \textbf{Typical metrics} & \textbf{Relevant standards/regulatory frameworks} \\
    \midrule
    Mechanical vehicle & Physical performance, durability & Classical engineering \\
    Early ADAS & Detection accuracy, false-alarm rate; modular performance & ISO~26262 (strict) \\
    Modern ADS & Time-to-collision, disengagement rate; scenario pass/fail & ISO~26262 + ISO~21448 (SOTIF) \\
    Data-driven ADS & Explainability, transparency, uncertainty; statistical assurance at scale & ISO/PAS~8800, UL~4600, UNECE R157 \\
    \bottomrule
  \end{tabular}
\end{table}

When the vehicle is the reasoning agent, it operates in an open world rather than a bounded test envelope \citep{Leahy2026OpenWorld}; and its behaviour is defined implicitly by training data. The relevant disciplines become statistical learning theory, human-factors science and uncertainty quantification, with explainable AI as the emerging bridge between them \citep{iso21448,iso8800}. The difference is that they play a central role in establishing safety, while many existing industry processes, standards and skills were developed primarily around mechanical systems.

Table~\ref{tab:metrics} summarises how the metrics used to make that argument have evolved. Early driver-assistance functions were validated on detection accuracy and false-alarm rate, component by component, under strict ISO~26262 discipline. Modern automated-driving generation added behavioural metrics such as time-to-collision and disengagement rate. The data-driven generation adds transparency, explainability and, as its ultimate goal, statistical assurance at scale---the demonstration, to authorities, the public and one's own management, that a learned system is acceptably safe across the situations it will encounter. New standards appear to support this \citep{iso8800,ul4600}, but they are being written while the systems they govern are already on the road; ISO/PAS 8800, published in 2024, is already undergoing revision.

\subsection{Case Study: Intelligent Speed Assistance}
\label{sec:validation:isa}

The difficulty is easiest to appreciate for a function that looks trivial. Intelligent Speed Assistance (ISA) informs the driver of the prevailing speed limit and warns, or gently resists, when it is exceeded; it has been mandatory in new vehicles in the European Union since 2024 \citep{eu2019gsr}. The nominal design combines a digital map that may carry speed-limit attributes, a forward camera that reads speed signs, and GNSS positioning to localise the vehicle on the correct road. 

In operation, the map is incomplete or out of date; signs are missing, obsolete but never removed, duplicated, occluded, or valid only for a category of vehicle or a time of year; and GNSS cannot always resolve which of two adjacent roads (a motorway and its parallel service road) the vehicle is on. The system must determine if the vehicle is travelling on the expected road, if a detected sign applies to that road or to an adjoining road, and if the sign applies to the vehicle class.

\begin{figure}[pos=htbp]
  \centering
  \includegraphics[trim={0 1.7cm 0 1.7cm}, clip, width=0.8\linewidth]{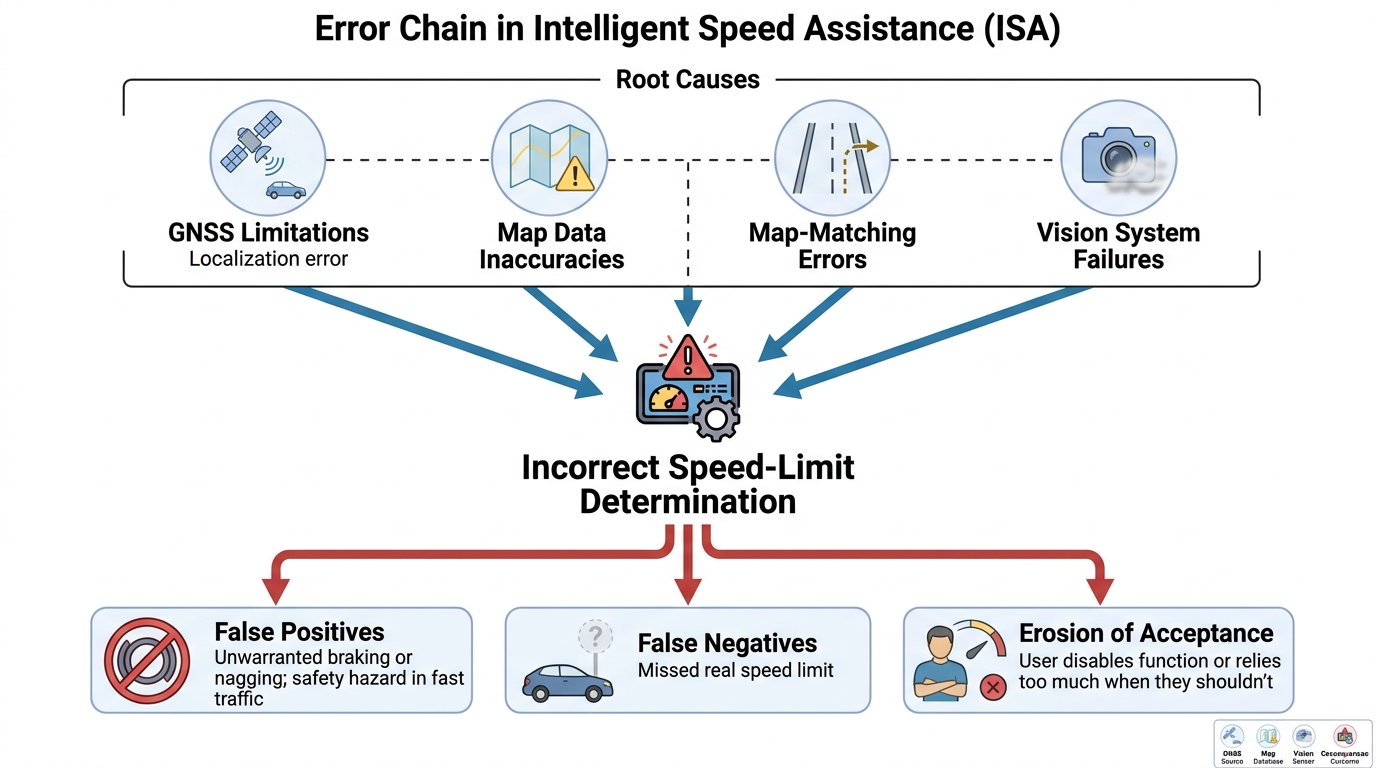}
  \caption{Error chain for Intelligent Speed Assistance.}
  \label{fig:isa}
\end{figure}

Fig.~\ref{fig:isa} traces the error chain. Root causes in positioning, mapping, map-matching, and vision propagate to an incorrect speed-limit determination, which manifests in three ways. False positives cause the vehicle to brake, or the driver to be nagged, for a limit that does not apply -- a safety hazard in its own right in fast traffic \citep{HUANG2026115230}. False negatives cause a real limit to be missed. And both erode acceptance: users who find the function unreliable switch it off, while users who find it mostly reliable begin to trust it where they should not.

Validating even this function requires the full stack: localisation, map, map-matching, and vision, each validated in isolation, then the classical integration process, then a statistical assessment of the fused behaviour. A validation campaign completed in France does not transfer to Germany, whose sign conventions differ, nor to the village whose mayor has posted a 15~km/h limit, nor to the urban core where a large limit sign carries a small supplementary panel that only residents know applies to delivery vehicles. Locality is not an edge case; it is the ordinary condition of the road network. Locality covers both local rules and signage and regional driving culture, such as following distances and gap acceptance.

\subsection{The Validation Toolbox and the Billion-Mile Paradox}
\label{sec:validation:toolbox}

Software-in-the-loop testing can exercise millions of virtual kilometres but represents the operating environment only as well as its models do. Hardware-in-the-loop brings the real electronic control units into the loop; vehicle-in-the-loop adds the real vehicle; driver-in-the-loop includes the human. Beyond that lie proving grounds (where consumer-rating protocols such as Euro~NCAP impose increasingly elaborate, high-fidelity test procedures \citep{EuroNCAP2026Safety}) and finally naturalistic driving on public roads. As one climbs the ladder the fidelity rises and the number of kilometres that can be driven falls, because each kilometre becomes more expensive and physically harder to stage.

\begin{figure}[pos=htbp]
  \centering
  \includegraphics[trim={0 0cm 0 2.2cm}, clip, width=0.8\linewidth]{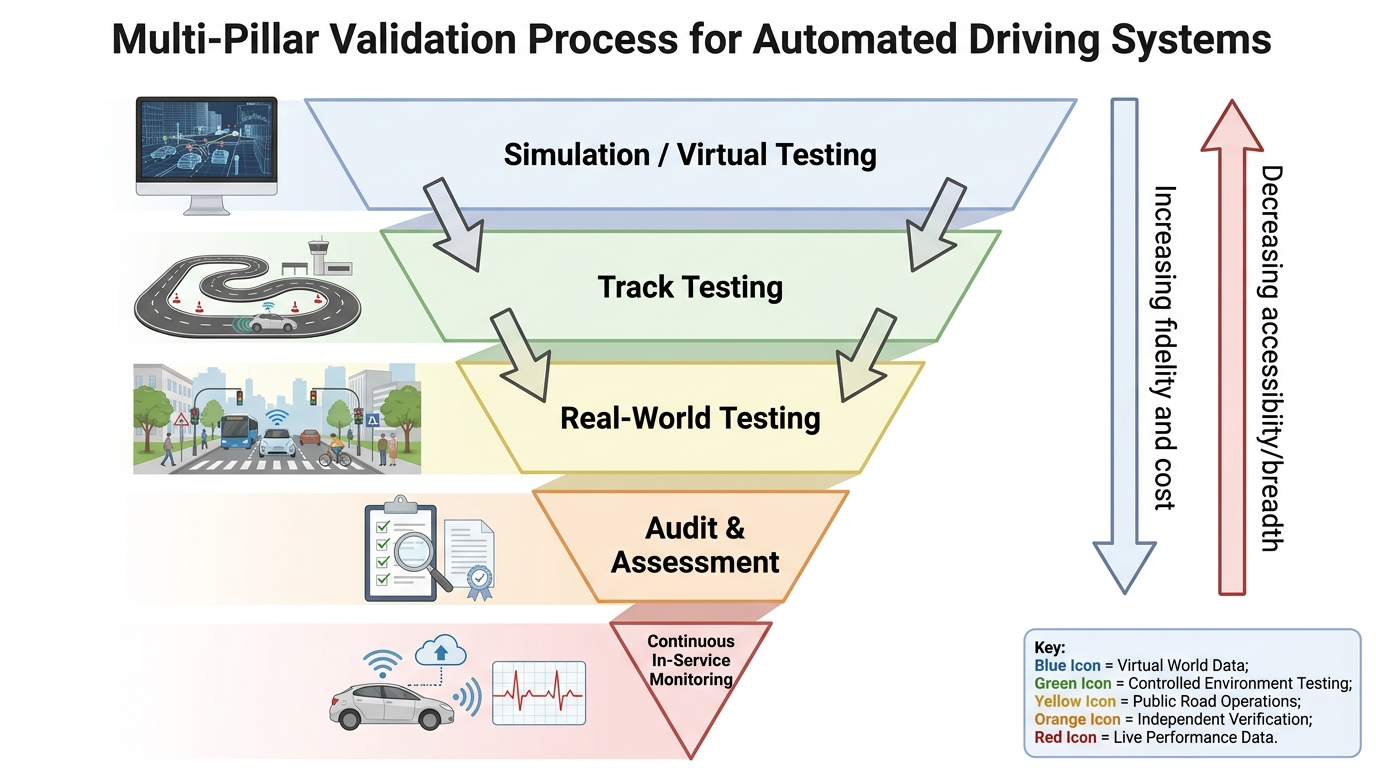}
  \caption{Multi-pillar validation.}
  \label{fig:pyramid}
\end{figure}

Statistical validation that an automated system is safer than a human at the level of fatal crashes requires on the order of hundreds of millions to billions of kilometres of representative driving, an amount no test fleet can accumulate \citep{kalra2016driving,wachenfeld2016release}. Scenario-based testing was introduced to avoid this \citep{riedmaier2020survey}, but it carries its own combinatorics: a single intersection crossing admits a large space of manoeuvres by other agents, each combined with initial conditions, weather and lighting, and it is far from obvious how many of those combinations must be exercised before the function can be declared to operate correctly. We refer to this as the billion-mile paradox: the closer a method comes to the real environment, the less of the required volume it can deliver.

Fig.~\ref{fig:pyramid} depicts a multi-pillar structure: simulation and virtual testing at the base; track testing; real-world testing; independent audit and assessment; and, increasingly, \emph{in-service monitoring}. The last pillar has changed the most: because vehicles are continuously connected, false alarms, incidents and disengagements can be uploaded and analysed across the deployed fleet, subject to privacy regulation. Validation need no longer end at type approval but can continue throughout the product's lifecycle. Initiatives such as White Box Autonomy \citep{zheng2026whitebox} and OpenPAV \citep{zhou2026openpav} can play a valuable role in supporting this continuous, evidence-based evaluation.
This matches the fleet-driven edge-case mining loop that end-to-end developers use for training (Section~\ref{sec:background:training}); the knowledge gap, taken up in Section~\ref{sec:discussion}, is who has access to that data.

\subsection{The Operational Design Domain as a Bounding Device}
\label{sec:validation:odd}

The operational design domain (ODD) has become an integral part of every validation process \citep{iso34503,thorn2018framework}. The idea is older than its name (robotics called it the workspace) and its function is to bound the problem. Environmental conditions, geography, infrastructure type, traffic composition and communications availability (including GNSS coverage) are specified in advance, and the vehicle is validated for that bounded space.

The ODD defines a theoretical envelope, whereas the environment encountered in real-world operation may include conditions outside that envelope, and outliers are, by construction, where trouble occurs. Bounding only works if the system can detect when it is leaving its limits. If the ODD can be monitored, the vehicle can degrade gracefully or hand over; if it cannot, the vehicle will at some point be operating outside its validated capability with no one aware of it, placing every road user around it at risk. 

In SAE Level~3 the system can request that the human take over the control of the vehicle. The driver's mind must then re-enter a situation almost immediately, without knowledge of the preconditions that led the vehicle there, an evaluation problem for which no satisfactory validation method yet exists. In Level~4 the question moves to the boundary: can the ODD be enforced, and what is the evidence that exit from it is always detected?
Supervised end-to-end systems such as Tesla’s FSD (Supervised) are Level 2 driver support features. The system performs sustained lateral and longitudinal control and much of the object and event detection across whole trips, while the driver must supervise continually.

\subsection{Case Study: In-Service Evaluation as Validation Evidence}
\label{sec:evaluation}

To illustrate how the validation methodology described above can be applied beyond pre-deployment testing, we consider the objective evaluation of ADAS functions during naturalistic driving. In-service evaluation provides empirical evidence of how these systems perform under real-world conditions and can reveal differences between implementations that may not be captured by track-based or subjective assessments.
 
Lane support functions, Lane Departure Warning (LDW) and Lane Keeping Assist (LKA), are a natural starting point. They are fitted to most new vehicles and their behaviour can be largely described by one quantity: the lateral position of the vehicle relative to the lane boundary. That quantity is easy to measure on a test track and hard to measure anywhere else.

Various lane support test protocols exist in the United States, Europe, Japan, and at the international level \citep{EuroNCAP2026Safety, nhtsa2013ldw, iso11270}. Most specify controlled departures on straight and curved sections of a closed track. The usual metrics are distance to lane edge (DTLE), measured at the start of the warning or at the maximum excursion, and, for LKA, lateral acceleration or jerk. In the Euro~NCAP LKA test, for example, the vehicle drifts out of the lane at 72~km/h and passes if DTLE stays within a set limit \citep{EuroNCAP2026Safety}.
 
Research setups, such as those described by \citet{Ma2020LKAS, Nguyen2021LDWLKA}, typically use RTK-corrected GNSS and an inertial unit supported by a base station, surveyed lane markings, and access to the vehicle CAN bus for steering, pedal, and wheel-speed signals. These methods measure the right things, but only in short, controlled sections of the road, with infrastructure that is costly to set up and CAN access that manufacturers are less and less willing to provide. They say little about how a function behaves in daily driving.

An alternative, currently being developed by the Australian Centre for Robotics \citep{Fitzgerald2025ADAS}, is a self-contained measurement kit that can be moved between cars, does not require infrastructure or CAN access, and can be used in normal traffic. The quantities remain the same; only the way they are obtained changes (Table~\ref{tab:sensing}). Cameras mounted low on the doors see the tyre and the nearby lane marking, and hence the calculation of the lateral position is possible. A standard GNSS receiver provides speed, an IMU gives acceleration, and an in-cabin camera and microphone aimed at the dashboard record driver inputs, ADAS status, and warning sounds. Data are logged during the drive and processed offline.

\begin{table}[pos=htbp]
  \caption{Sensing strategy of the portable system compared with existing approaches.}
  \label{tab:sensing}
  \centering
  \small
  \begin{tabular}{@{}p{3.0cm}p{4.0cm}p{5.4cm}@{}}
    \toprule
    \textbf{Quantity} & \textbf{Existing approaches} & \textbf{Portable system} \\
    \midrule
    Lateral position & Surveyed stripes + RTK GNSS/IMU & Door-mounted cameras, fiducial calibration, SAM3 lane detection \\
    Speed & CAN wheel speeds / RTK & Consumer GNSS \\
    Acceleration & Roof IMU + base station & Compact IMU \\
    Control inputs, alerts & Proprietary CAN + in-cabin camera & In-cabin camera on dashboard + microphone \\
    Infrastructure & Base station, survey, closed track & None \\
    Installation & Roof rack, per-vehicle fitting & Suction mounts, 5--10~min \\
    \bottomrule
  \end{tabular}
\end{table}

By calibrating the system, we can convert image coordinates to metric distances. A board with fiducial markers is placed on the ground next to each wheel and the homography between the image and the road plane is estimated, so that any detected lane-marking point can be converted into a lateral distance from the tyre. Lane markings can be segmented with a general-purpose foundation model such as SAM3 \citep{sam3}. Dashed lines come out as separate segments, so a temporal filter is needed to join them into one boundary and smooth the estimate from frame to frame \citep{Wang2026SideViewLane}. Rain and dirt on the lenses reduce the detection quality, but in practice the boundary can still be tracked.

\begin{figure}[pos=htbp]
  \centering
  \includegraphics[trim={0 0cm 0 0cm}, clip, width=0.8\linewidth]{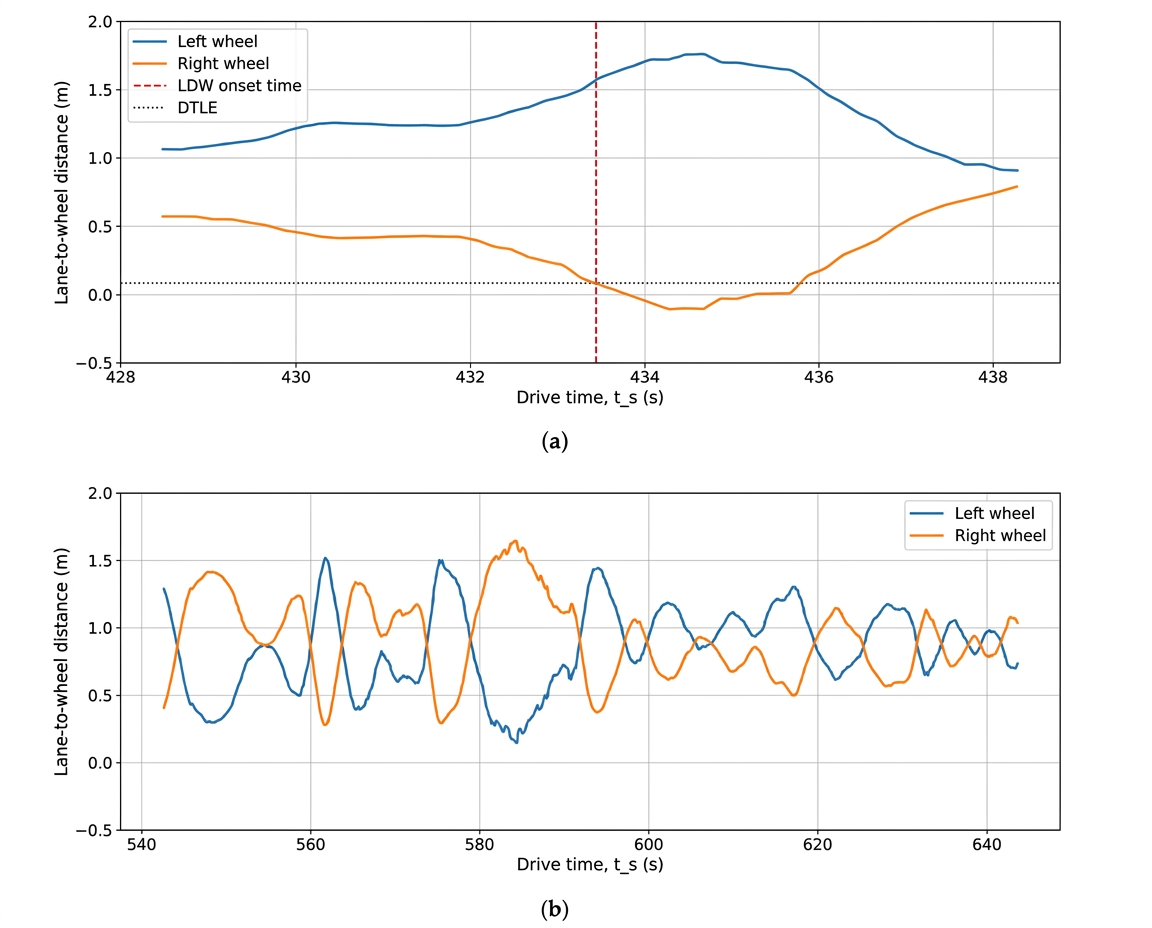}
  \caption{Lateral distance estimates during (\textbf{a}) an LDW event and (\textbf{b}) a 100-second LKA segment for two production cars. Preliminary data from the authors' ongoing work.
  }
  \label{fig:lateral}
\end{figure}

The output contains two time series with the distance from the left and right wheels to their lane markings (Fig.~\ref{fig:lateral}).
For LDW, the warning onset is timestamped from the cabin audio and the lateral distance at that moment is read from the signal. This is the same DTLE used in track protocols, but measured on real roads, at the speeds and road geometries drivers actually encounter. For LKA, what matters most is how the vehicle moves within the lane over time. A good system keeps the vehicle near the centre of the lane. The amplitude, frequency, and asymmetry of the oscillation together with lateral acceleration and approach speed, describe the quality of its perception and control. How these measures are combined into one score is still unclear, as manufacturers may give different importance to comfort, accuracy, and intervention timing. The vehicle’s approach speed, as used by Euro NCAP, can also determine which events are included. The measurement leaves the choice of metric open, while making a range of metrics computable from naturalistic data.

% \begin{figure}[pos=htbp]
%     \centering
%     \subfloat[\centering]{\includegraphics[width=0.8\textwidth]{Figures/ldw__recording_20260115T134749__seg_000_000428440_000438440.png}}
%     \vfill
%     \subfloat[\centering]{\includegraphics[width=0.8\textwidth]{Figures/lka__recording_20260224T111642__seg_002_000542617_000643706.png}}
%     \caption{Lateral distance estimates during (\textbf{a}) an LDW event and (\textbf{b}) a 100-second LKA segment for two production cars.}
%     \label{fig:lateral}
% \end{figure}

Lane-support functions are often rated by expert drivers or in user studies, and these ratings depend on the driver's familiarity with the vehicle, expectations, and personal preference. In a preliminary comparison of two production vehicles from ongoing, objective measurements of this kind ranked the two lane-support implementations in the same order as a structured driver study. With only two vehicles, this shows that the method can discriminate between implementations; it does not establish how large such differences are in general.
 
A similar approach can be applied to other functions, such as adaptive cruise control. More broadly, it suggests that the in-service monitoring identified in Section~\ref{sec:validation} as a growing part of validation can be applied to Level~2 functions on production cars, without the manufacturer's involvement, and can reveal differences between implementations that track tests and subjective ratings may miss.

%%%%%%

\subsection{Trends and Open Problems}
\label{sec:validation:trends}

We identify five trends, each with a corresponding gap:

\textit{From component reliability to system-level scenarios.}
Evidence is shifting from component-level reliability figures to
scenario-based assessment of whole-system behaviour. The gap is that
scenario coverage is undefined for open-world operation.

\textit{Definitions clarified, benchmarks still missing.}
The 2026 revision of SAE J3016 \citep{SAEJ3016_2026} rejects labels such as “L2+”: levels are mutually exclusive integers, and differences within a level are described with qualifiers such as “hands-off capable”. Supervised end-to-end systems are therefore Level 2 features, set apart by the breadth of their ODD rather than their level. The taxonomy imposes no performance requirements, however, and there are still no common benchmarks across levels.

\textit{Rare-event statistics.}
Statistical demonstration of safety under rare events is still badly underestimated, and the lessons learned in validating L1 and L2 functions do not transfer directly to L4 and L5.

\textit{Explainability at failure.}
When an end-to-end system misbehaves it is hard to establish what happened. The auxiliary outputs, attention maps and language rationales noted in Section~\ref{sec:background:consequences} help, but their faithfulness is unverified.

\textit{Over-the-air updates.}
Software-defined vehicles receive frequent updates. Each update must be validated for the function it changes and for regression across everything else in the vehicle, and it is not settled if that requires full re-simulation, partial re-testing, or something not yet defined.

These gaps call for renewed verification and validation research, a need already recognised by ongoing standardisation such as IEEE P2817 \citep{IEEEP2817}, whose development has shown how rapidly verification complexity grows for autonomous systems: hybrid validation combining the pillars above; renewed interest in formal verification where applicable; advanced scenario generation to synthesise edge cases not yet observed in fleet data; and a return to statistical design of experiments. Three dimensions of uncertainty need to be characterised (perception, user behaviour, and the boundaries within which the system moves) so that limits are known in advance. And finally, the tools themselves need to be certified. Repeated simulations of the same scenario may produce inconsistent results, raising concerns about the reliability of simulation-based validation. The strength of any validation argument depends not only on the consistency of the simulator but also on its fidelity to real-world operating conditions. This includes whether the simulator adequately captures factors such as sensor degradation, thermal cycling, mechanical vibrations, and component ageing, all of which can affect system performance and safety. Verification in open-world conditions is itself recognised as a grand challenge for autonomous systems \citep{Leahy2026OpenWorld}.

\subsection{Should the Vehicle Adapt to the Road, or the Road to the Vehicle?}
\label{sec:validation:infra}

Much of the ISA difficulty arises from infrastructure that predates the technology: signs that are missing, ambiguous or local. Drivers experience the same issue (anyone who has missed a sign in an unfamiliar area has driven without knowing the speed limit) and at least one jurisdiction solved it a decade ago not by improving the vehicle but by publishing a geo-referenced database of every posted limit, which a smartphone application then used to advise the driver \citep{NSWGovernment2024SpeedAdviser, EuropeanCommission2021ISA}. The argument is that if speed-limit provision were treated as a data service maintained by the road authority, the vehicle's vision system would become a cross-check rather than the primary source, and the validation burden would fall dramatically.

Europe has tried to make road signs more consistent, but local rules and unusual signs still vary widely. These local details, especially in older cities, are difficult for a central database to capture and keep up to date. This challenge reflects a broader need for AV-friendly road infrastructure \citep{Zheng2026Conversation}: rather than expecting automated vehicles to adapt to every limitation of infrastructure designed primarily for human drivers, road authorities should also consider how physical and digital infrastructure can better accommodate automated driving. This does not necessarily require extensive infrastructure upgrades; clearer road markings, more consistent signage and reliable digital road information could substantially improve the operating environment for both automated vehicles and human drivers. The resulting research and policy questions extend beyond vehicle technology: which aspects of the operational design domain should be supported by authoritative infrastructure data, who should maintain them, and how should vehicle validation account for their availability and reliability? We return to this in Section~\ref{sec:discussion}.

\section{What Is Good Driving? From Crash Statistics to Behavioural Assurance}
\label{sec:gooddriving}
 
The preceding sections considered how automated-driving behaviour can be understood and how evidence for its safety can be constructed. But, against what standard should an automated vehicle’s driving be judged?
 
\subsection{Seeing What the Vehicle Thinks}
\label{sec:gooddriving:insight}
 
Classical driver-behaviour research has a well-known limitation \citep{Michon1985}: experiments reveal what a driver does in a scenario, but not how the driver's brain interpreted the information available and how that interpretation shaped the decision. For human drivers, the gap is tolerated because human cognition is broadly shared and understood. For automated vehicles it is not tolerable, because the vehicle's interpretation can fail in ways no human's would \citep{7823109}.
 
The fatal collision between an Uber test vehicle and Elaine Herzberg in Tempe, Arizona, in March 2018 left a mark. The investigation found that the automated driving system detected her several seconds before impact, but repeatedly changed its classification of her (between vehicle, bicycle and other) as she walked a bicycle across the road away from a crossing, and as a result failed to predict her path; the safety operator, who should have intervened, was not watching the road \citep{ntsb2019uber}. We argue that, had the vehicle's classification uncertainty been observable as it drove rather than reconstructed after the event, the testing and deployment of that system might have been approached with considerably more caution. The requirement that follows is general: assurance of an automated vehicle needs insight into what it perceives, predicts, and intends, not only into what it does.
 
Robotaxis, including one built on an end-to-end driving stack, began piloting in London in 2026 with the stated aim of moving to driverless operation \citep{wayve2026london}, under the regulatory framework of the Automated Vehicles Act 2024 \citep{avact2024}.
 
\subsection{``Careful and Competent'': A Legal Standard Without a Metric}
\label{sec:gooddriving:standard}
 
Section~2 of the Automated Vehicles Act \citep{avact2024} requires a statement of safety principles framed so that authorised automated vehicles achieve a level of safety equivalent to, or higher than, that of careful and competent human drivers, and so that road safety in Great Britain is better as a result of their use. What it does not say is how a developer could prove that either condition has been met.
 
The obvious approach is statistical: show that the automated fleet's rate of death and serious injury is lower than that of human drivers. Table~\ref{tab:benchmarks} shows why this is less straightforward than it appears. In 2024 Great Britain recorded 1,602 road deaths over roughly 340 billion vehicle miles \citep{dft2025ras}, about 212 million miles per fatality. But that figure includes motorcyclists, whose risk is far higher than that of car occupants, and drivers who were fatigued, distracted or intoxicated, the very drivers who are not careful and competent. The same calculation for the United States yields roughly 83 million miles per fatality \citep{nhtsa2025fars}, about two and a half times the British fatality rate, so a demonstration that a vehicle improves road safety in the United States does not prove that it would do so in Great Britain.
Restricting the British calculation to incidents involving cars and removing fatalities in which human error was recorded as a contributory factor gives a benchmark that represents careful and competent driving: approximately 2.5 billion miles per fatality.
 
\begin{table}[pos=htbp]
  \caption{Fatality-rate benchmarks for ``careful and competent''
  driving, computed from \citet{dft2025ras,nhtsa2025fars}}
  \label{tab:benchmarks}
  \centering
  \small
  \begin{tabular}{@{}p{7.0cm}r@{}}
    \toprule
    \textbf{Population (2024)} & \textbf{Miles per fatality} \\
    \midrule
    United States, all road users & $\approx 83$~million \\
    Great Britain, all road users & $\approx 212$~million \\
    Great Britain, car-involved, excluding human-error contributory factors & $\approx 2.5$~billion \\
    \bottomrule
  \end{tabular}\\[2pt]
\end{table}
 
Those numbers are higher than all the automated driving completed by all developers so far. It also connects to the arithmetic of Section~\ref{sec:validation:toolbox}. Demonstrating a fatality-rate improvement statistically requires hundreds of millions to hundreds of billions of miles \citep{kalra2016driving}. Even a fleet of 1,000 vehicles averaging 30 mph continuously would require roughly 380 years to accumulate 100 billion fleet miles. Fatality statistics by themselves therefore cannot support authorisation. %Statistical demonstration at the fatality level is not a route to authorisation; other forms of evidence are required.
 
\subsection{Statistics Are Not How the Public Judges Safety}
\label{sec:gooddriving:public}
 
The British experience with smart motorways (where the hard shoulder was converted to a running lane and technology added to detect stopped vehicles) offers a useful comparison. Statistical evidence shows that these roads perform as well as or better than other motorway types \citep{dft2021smart}, yet public opposition has been sustained, driven by a visceral fear of stopping in a live lane and by the small number of tragic crashes in which this happened. A safety case based only on overall statistics can quickly lose public trust after a serious incident. The same happened to automated vehicles in October 2023, when a robotaxi in San Francisco struck a pedestrian who had been thrown into its path by another car and then dragged her about 6~m; the operator's incomplete reporting of the event to regulators compounded the damage, and its driverless operations ended soon after \citep{nhtsa2024cruise}.
 
Interviews and workshops with members of the public show that people do not reason about automated vehicle safety in the industry's terms. They do not focus on miles driven, crash rates, disengagements, or the number of people killed or seriously injured. They talk about behaviour: whether the vehicle detects hazards appropriately, is aware of other road users, drives considerately and courteously, and behaves predictably. These are the same qualities the drivers interviewed in Section~\ref{sec:human:hmi} cited when they said that they trusted the vehicle because they knew what it would do. %Statistics remain the foundation of the industry's understanding; public acceptance is built on behaviour.
 
Following that research, the government's statement of safety principles was reframed in behavioural terms: the draft published for consultation in June 2026 sets out ten principles grounded in behaviours associated with careful and competent driving, which developers will be required to show their vehicles exhibit \citep{dft2026sosp}. Under the draft, self-driving vehicles should:

\begin{enumerate}
  \item use the road in accordance with traffic laws and the Highway Code;
  \item maintain control over the vehicle at all times;
  \item predict, detect and respond to hazards proactively;
  \item adapt driving to road and weather conditions;
  \item behave in a predictable manner;
  \item improve road safety in Great Britain for all;
  \item interact safely with all road users, particularly vulnerable road
        users;
  \item interact safely with emergency services;
  \item operate autonomously only within the limits under which they are
        specifically designed to function; and
  \item be designed to take account of the specificities of the territory
        where they will operate.
\end{enumerate}

Only one of the ten, the sixth, is an outcome measured in aggregate statistics; the rest describe how the vehicle should drive. They map closely onto the gaps identified in this paper. Proactive hazard prediction (3) is the subject of Section~\ref{sec:prediction}; predictability (5) is what the supervising drivers of Section~\ref{sec:human:hmi} valued most; operating only within designed limits (9) is the operational-design-domain monitoring issue of Section~\ref{sec:validation:odd}; and accounting for the specificities of the territory (10) is the locality obstacle illustrated by intelligent speed assistance in Section~\ref{sec:validation:isa}. The principles do not specify how a developer would validate any of them, or how a regulator would check. Recent cases show why behavioural evidence matters, a robotaxi fleet was recalled after its vehicles repeatedly passed stopped school buses \citep{waymo2025schoolbus}, and a supervised end-to-end system is under investigation for running red lights and entering opposing lanes \citep{nhtsa2025pe25012}. Neither pattern would be visible in fatality statistics, but both breach the first principle.
 
There is a further asymmetry. A passenger who gets into a car with a human driver trusts implicitly that the driver does not want to be hurt, does not want to hurt them, does not want to damage the vehicle or anyone's property, broadly intends to follow the rules of the road, and knows that breaking them carries personal consequences. Those instincts are the product of evolution and socialisation. An automated vehicle does not have them by default, and nothing in its behaviour signals that it does. The missing instincts must be replaced by data showing that the vehicle behaves as if it had them.
 
\subsection{Digital Commentary Driving: A Proposal for Discussion}
\label{sec:gooddriving:dcd}
 
One proposal, developed for the British Standards Institution, is \emph{Digital Commentary Driving} (DCD) \citep{bsi2021dcd}. It was derived by comparing how human drivers are assessed, how automated vehicle developers assess their own systems, and how mobile robotics assesses autonomous platforms. Its name comes from commentary driving, a technique used to train and test human drivers in which the driver verbalises what they observe, what they expect and what they intend to do \citep{omeiza2022spoken}; examiners use it to judge the quality of the driver's hazard perception and planning, not merely the outcome.

\begin{figure}[pos=htbp]
  \centering
  \includegraphics[trim={0 3cm 0 3cm}, clip, width=0.8\linewidth]{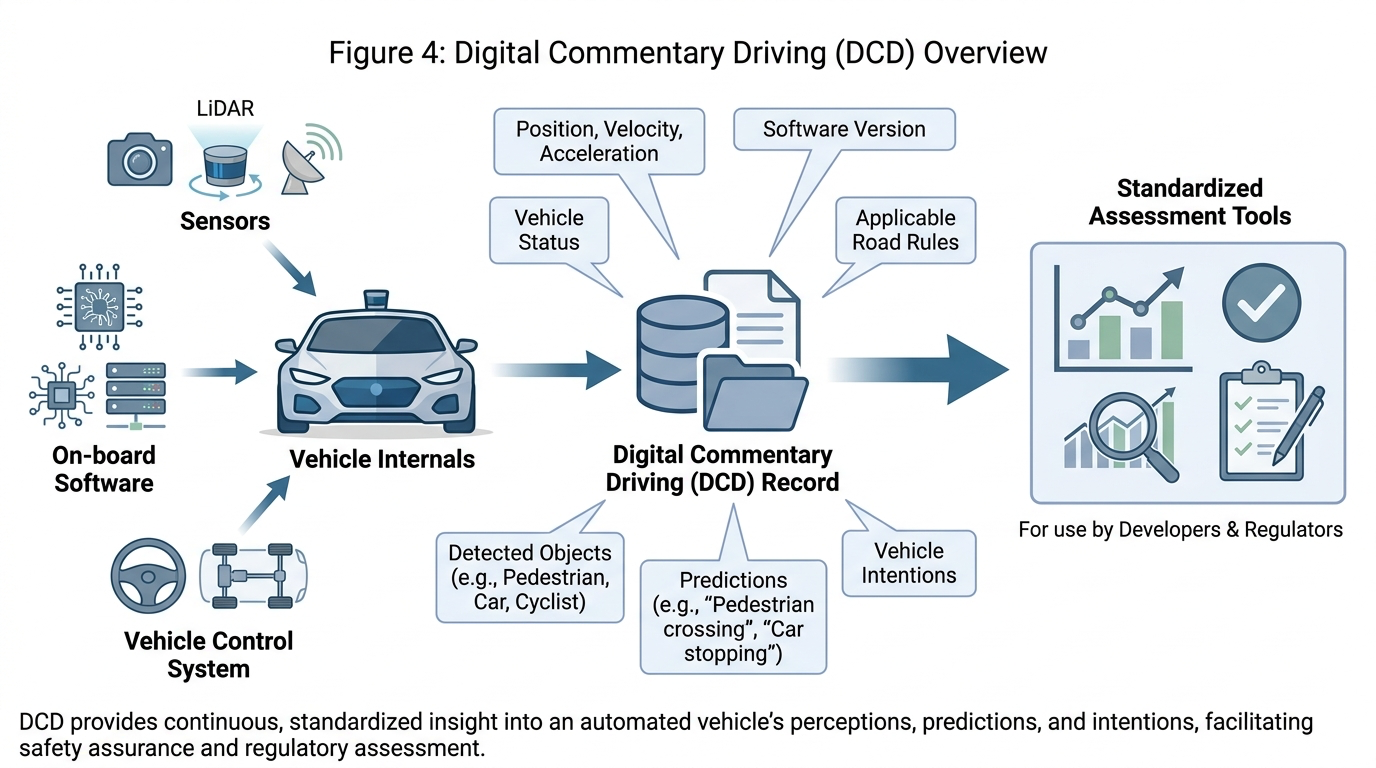}
  \caption{Digital Commentary Driving overview.}
  \label{fig:dcd}
\end{figure}
 
DCD is the machine equivalent: a standardised data format in which the vehicle records, continuously or on trigger, what it is doing, what it perceives, what it predicts, and what it intends next. The content includes position, velocity and acceleration; vehicle status and software version; the road rules the vehicle believes apply; the objects it has detected; and their expected behaviour, such as a pedestrian preparing to cross or a vehicle approaching from a side road that may not stop in time (Fig.~\ref{fig:dcd}). With that information available from every drive in a common format, standardised tools can assess whether the vehicle is driving at a careful and competent level.

Four properties make the proposal relevant to the end-to-end transition described in this paper.
 
\textit{Architecture-agnostic.}
The DCD record sits outside the developer's intellectual-property boundary. Regardless of the sensing technology used (lidar, radar, cameras, ultrasound or V2X) and the software architecture (rule-based or end-to-end neural network) the vehicle must be able to access and report the information required for safe driving. This gives a solution to the concern raised in Section~\ref{sec:background:consequences} that end-to-end systems lack the intermediate representations on which validation has traditionally relied: DCD specifies the representations an assessor needs without dictating how they are produced.
 
\textit{Small, and usable everywhere.}
The data are deliberately simple (no video, no point clouds) and compress to the order of kilobytes per hour of driving. The same format applies in simulation, on test tracks and on public roads; it can be analysed for portions of a drive or whole drives; and supports comparison between manufacturers, or between successive software versions of the same vehicle. The last capability speaks to the over-the-air regression issue of Section~\ref{sec:validation:trends}.
 
\textit{A tool for regulators.}
The intended recipient is the regulator. Industry has resisted this kind of access to what the vehicle sees and does, but the data would be held confidentially, not published, and their small size and standard format would let regulators apply standardised analysis tools instead of relying on the developer's own interpretation. This is how DCD would address the imbalance in resources and capability between the companies developing automated vehicles and the authorities charged with judging whether they are safe, the same imbalance raised in Section~\ref{sec:discussion:data}. For it to work, the approach must be adopted broadly rather than imposed in one jurisdiction as a restriction
others do not share.
 
\textit{Data, not narration.} Some developers provide natural-language descriptions of what their vehicles are doing. We regard these as interesting and valuable for public understanding but insufficient for assurance: converting the vehicle's operation into language is inherently lossy, and assurance needs the underlying data. This sharpens the faithfulness concern of Section~\ref{sec:background:consequences}: even a faithful rationale is a compressed summary, and an assessor needs what it was compressed from.
 
\subsection{Next Steps and the Read-Across to Human Driving}
\label{sec:gooddriving:next}
 
DCD is a proposal, and the next step is empirical. Data must be collected in the standardised format at scale and used to answer the questions on which the concept depends: can careful and competent driving actually be assessed from such data; what thresholds separate acceptable from unacceptable behaviour; how should those thresholds vary with the context of operation; and can the analysis be packaged in tools that non-experts, including regulators, can use?
 
What it means to be a good driver is, for automated vehicles, still a work in progress. We argue that the bar should be set high, both to deliver the safety benefits that have been promised for automated vehicles for many years and to meet the public expectations those promises have created. A high bar protects the whole industry: an incident involving the least careful developer damages the reputation of the most careful, so it is in every developer's interest to exclude those not doing an adequate job.

Beyond safety assurance, the evaluation of supervised end-to-end driving should consider its broader implications for transport systems and society. \citet{zheng2026seedsav} proposed SEEDS-AV, a multidimensional, multiscale and human-centred framework encompassing five dimensions: Safety, Equity, Efficiency, Disaster-readiness and Sustainability. The framework treats safety eligibility as a non-compensatory requirement while enabling separate assessments of safety performance and other dimensions, drawing on complementary evidence from simulation, controlled testing and real-world observations. Such comprehensive assessment frameworks are essential for defining and evaluating what constitutes good driving by automated vehicles.
 
Finally, the influence may run in both directions. Once a bar for good automated driving is defined and the data associated with it are understood, the same measures may change how human driving is assessed.
Many collisions involve behaviour that could obviously and easily have been different; what is learned about careful and competent driving in the automated context may help define, measure and improve it in people too, an outcome of value well beyond the automated vehicle itself.

%%%%
\subsection{Case Study: Behaviour Understanding for Assurance}
\label{sec:prediction}

A prediction layer matters for behavioural assurance because it can expose assumptions about surrounding road users that would otherwise remain latent within an end-to-end policy.

Perception in automated vehicles is very good at telling what is in the scene. It is much weaker at what human drivers do almost without thinking: entering a busy roundabout, expecting a pedestrian to step out from behind a parked van, sharing the road with a cyclist, or letting in a car that is merging aggressively from a slip road. These situations call for reasoning about what other road users are likely to do and for adjusting one's own driving accordingly. Behaviour prediction has a large literature, covering lane changes, cut-ins, pedestrian intention, and trajectory forecasting \citep{mozaffari2022review,rudenko2020human, rasouli2020autonomous}. Most of it treats prediction as an internal module of the driving stack. We argue that it is also an assistance function in its own right, useful for today's supervised vehicles, and still relevant inside an end-to-end architecture.
 
Good prediction can improve safety, traffic flow, and passenger comfort. It is also important to people outside the vehicle, who need to feel that the vehicle has seen them and will behave as they expect. That understanding is conveyed through the vehicle's motion and through external human--machine interfaces (eHMI) \citep{dey2020taming}, and it is a condition for wider acceptance.
Purely kinematic predictors only react once a manoeuvre has started, typically when lateral motion becomes visible. Human drivers anticipate earlier because they read the situation: a car that quickly closes on a slower vehicle, with a gap opening in the next lane, is likely to pull out. Adding this kind of contextual information extends the prediction horizon \citep{mozaffari2022review}.
 
One way to represent context is through a knowledge graph. Numerical quantities such as speeds, gaps, and time-to-collision with neighbouring vehicles are first mapped to linguistic categories (low, safe, critical, and so on), and the scene is then described as a set of triples linking the target vehicle to its neighbours. Knowledge-graph embeddings learned from these triples can feed a probabilistic classifier. In the highD motorway dataset \citep{krajewski2018highd}, this approach predicted lane changes several seconds before the vehicle crossed the lane marking \citep{manzour2024lane}. The same representation was used for pedestrian crossing intention, with features such as gaze towards the vehicle, walking pace, distance to the kerb and pedestrian type \citep{rasouli2019pie}. Related work has combined data and knowledge-driven methods for pedestrian prediction \citep{zhou2024pedestrian,halilaj2021knowledge}.
 
\begin{figure}[pos=htbp]
    \centering
    \includegraphics[width=\linewidth]{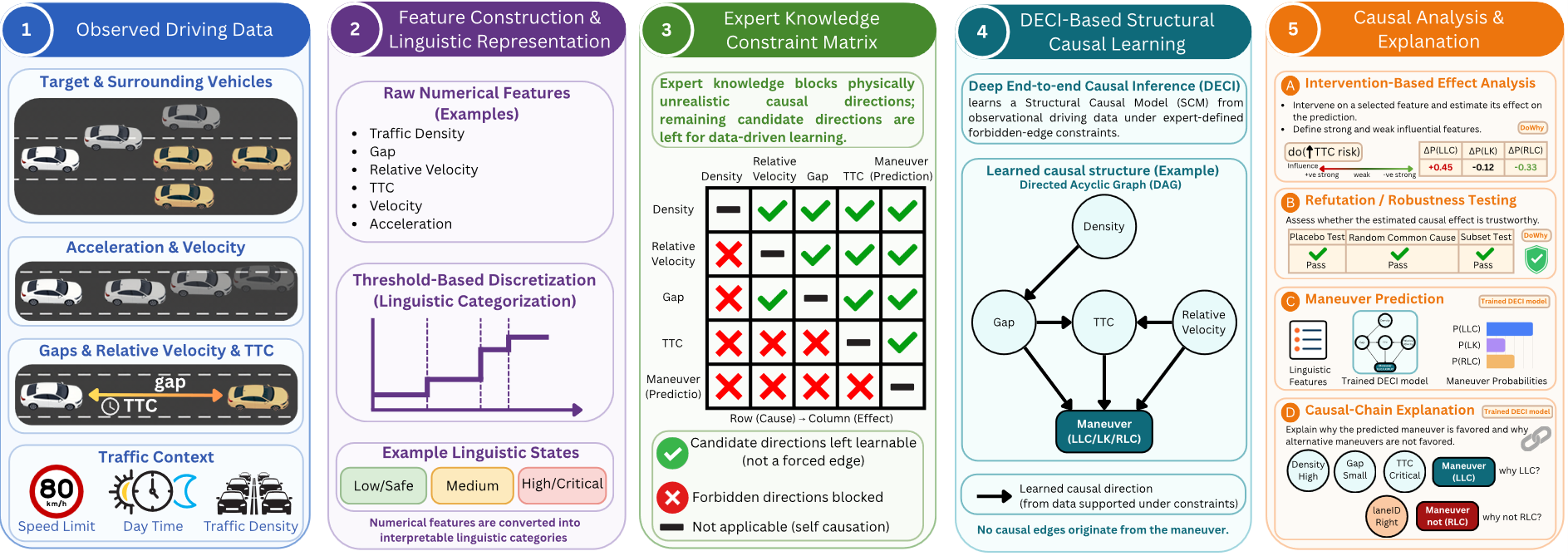}
    \caption{From correlation to cause: expert constraints, causal
    discovery and intervention analysis for lane-change prediction \citep{manzour2026correlation}.}
    \label{fig:causal_analisys}
\end{figure}

Most predictors treat their inputs as a flat set of independent features. Gap, relative speed, and time-to-collision are physically linked, traffic density influences speed rather than the reverse, and some variables are consequences of others. A feature such as time-to-collision may look like the main driver of a lane change, when it is really where upstream causes become visible. A recent study on lane-change prediction \citep{manzour2026correlation} provides an example of the approach (Fig.~\ref{fig:causal_analisys}). Expert knowledge first rules out causal directions that do not make physical sense, such as the relative speed that causes traffic density. Then a structural causal model is learned within these constraints using deep end-to-end causal inference \citep{geffner2022deep}, and intervention analysis measures how much each variable changes the predicted manoeuvre. The graph is checked with standard refutation tests such as placebo treatments and random common causes \citep{sharma2020dowhy}.
 
The learned relationships behave as expected of the driver. A vehicle drifting right becomes much more likely to change lanes to the right, and a large speed advantage over the vehicle ahead increases the probability of overtaking. The prediction accuracy is comparable to non-causal methods \citep{manzour2026correlation}. The causal view also helps with feature selection. If anticipation is measured at the start of the manoeuvre rather than at the lane-marking crossing, the lateral features do not carry information, and their removal does not change performance.
 
Prediction only helps if it changes what the vehicle does or what others understand about it. This can happen through the HMI, through the vehicle's motion, or both. 
The effect on safety could not be measured directly, but such gains relate to traffic flow and add to a growing body of work on how external displays shape pedestrian crossing decisions \citep{dey2020taming}.

Driving is shaped by local habits as well as by traffic law. Regions differ in how assertively gaps are taken and how much negotiation between road users is informal, so a system tuned in one region may not transfer to another. Two directions are promising: the use of ADAS to reinforce injunctive norms, that is, what people think others expect of them \citep{cialdini1990focus}, for example, by drawing attention to local rules and customs. The second direction aims at vehicles that can negotiate with other road users while respecting local conventions, using neuro-symbolic reasoning that combines traffic rules, social norms, and common sense. Common sense is hard to formalise, and it also varies from city to city.

\section{The Human in Supervised End-to-End Driving}
\label{sec:human}

%\subsection{The Unit of Safety Is the Human--AI System}
%\label{sec:human:unit}

The preceding section considered the vehicle as the focus of the safety assessment. Here we start from the observation that in supervised end-to-end driving the vehicle is not the thing whose safety matters. The AI performs most of the driving task; the human is responsible for supervising and intervening. The safety of the arrangement depends on how the two interact: trust, situational awareness, the driver's mental model of what the system can do, workload, and the quality of the interface between them \citep{endsley1995toward,lee2004trust}. Supervised end-to-end driving is, from the human's side, no different from what the field has been studying for years under the label of Level~2 and Level~3 automation, and the field's findings are not reassuring \citep{casner2016challenges}.

An end-to-end system holds a safe following distance without lapses, reacts to a sudden braking by a vehicle ahead faster than any human, monitors blind spots and vulnerable road users continuously through surround sensing, and is not subject to fatigue or distraction. Section~\ref{sec:background:training} described the reaction advantage: the policy samples the full scene every few tens of milliseconds, whereas a human needs on the order of half a second to perceive, glance away and return, by which time an urban scene may have changed materially. Where the machine is weaker is in ambiguity: unclear lane markings, construction zones that have changed since yesterday, intersections where right of way is not obvious. In such cases, the judgment of the driver is still required, and the driver has, by then, been idle for a long period of time. Another interesting aspect is how an automated vehicle can consider human passengers. Sometimes a human would prefer to change to a faster lane or, conversely, stay behind a specific vehicle regardless of traffic conditions. Can an end-to-end system consider these human preferences at the tactical and operational layers?

\subsection{Why Partial Automation Is the Hard Case}
\label{sec:human:hypovigilance}

The mechanism is well documented and \citep{bainbridge1983ironies} described in the terms the field has used since. Because the system performs almost all of the task, the driver's workload and effort fall. Reduced workload is not a benefit here: it reduces active engagement, and prolonged good performance produces \emph{hypovigilance}, in which the driver's capacity to respond quickly degrades because nothing has required a response. In parallel, sustained reliability produces \emph{over-trust}: drivers come to believe the system can handle everything, including the situations it cannot, and so intervene later than they should \citep{parasuraman1997humans,banks2018partially}. Crash investigations bear this out; in a fatal 2018 crash in Mountain View, California, a vehicle under partial automation steered into a highway barrier while the driver, who did not intervene, was probably distracted by a game on his phone \citep{ntsb2020mountainview}.
The objective is to calibrate driver trust to the actual capabilities and limitations of the automated system. Both excessive and insufficient trust can undermine appropriate use of automation \citep{OlaverriMonreal2020}. This is particularly relevant for supervised end-to-end driving, where prolonged successful operation may encourage drivers to assume that the system can also handle situations outside those they have previously experienced. Communicating system limitations and uncertainty is therefore as important as communicating its intended actions. 
Automotive manufacturers who have looked closely at this are reluctant to ship Level~3 at all, and consider Level~2 and Level~3 together the most dangerous region of the automation spectrum.

Moving eyes to the road after a warning takes roughly 0.8~s; but eyes on the road do not mean attention on the road, and even after a tactile alert has redirected gaze, the driver needs more time to assess the scene well enough to act \citep{zeeb2015what,gold2013take,eriksson2017takeover}.
Furthermore, the effects of distraction may persist even after the distracting activity has ended.  \citet{Winzer2017} demonstrated that drivers’ reaction times did not immediately return to their pre-distraction levels after completing a secondary task, indicating that recovery of attention requires additional time. The total time required to regain attention and control is strongly dependent on the non-driving task the driver was engaged in, and for some tasks regaining control in time is close to impossible. 
Experimental and real-world studies have further shown that the effects of automation on driver behaviour extend beyond delayed responses to takeover requests. Changes in visual attention, trust, and engagement can emerge during automated driving even when the system performs reliably. These findings highlight the need to evaluate driver supervision as a dynamic process in which the driver's readiness to intervene may change over time, rather than assuming a constant level of availability \citep{MoralesAlvarez2021}.

\begin{figure}[pos=h]
    \centering
    \includegraphics[trim={0 3cm 0 3cm}, clip, width=0.9\linewidth]{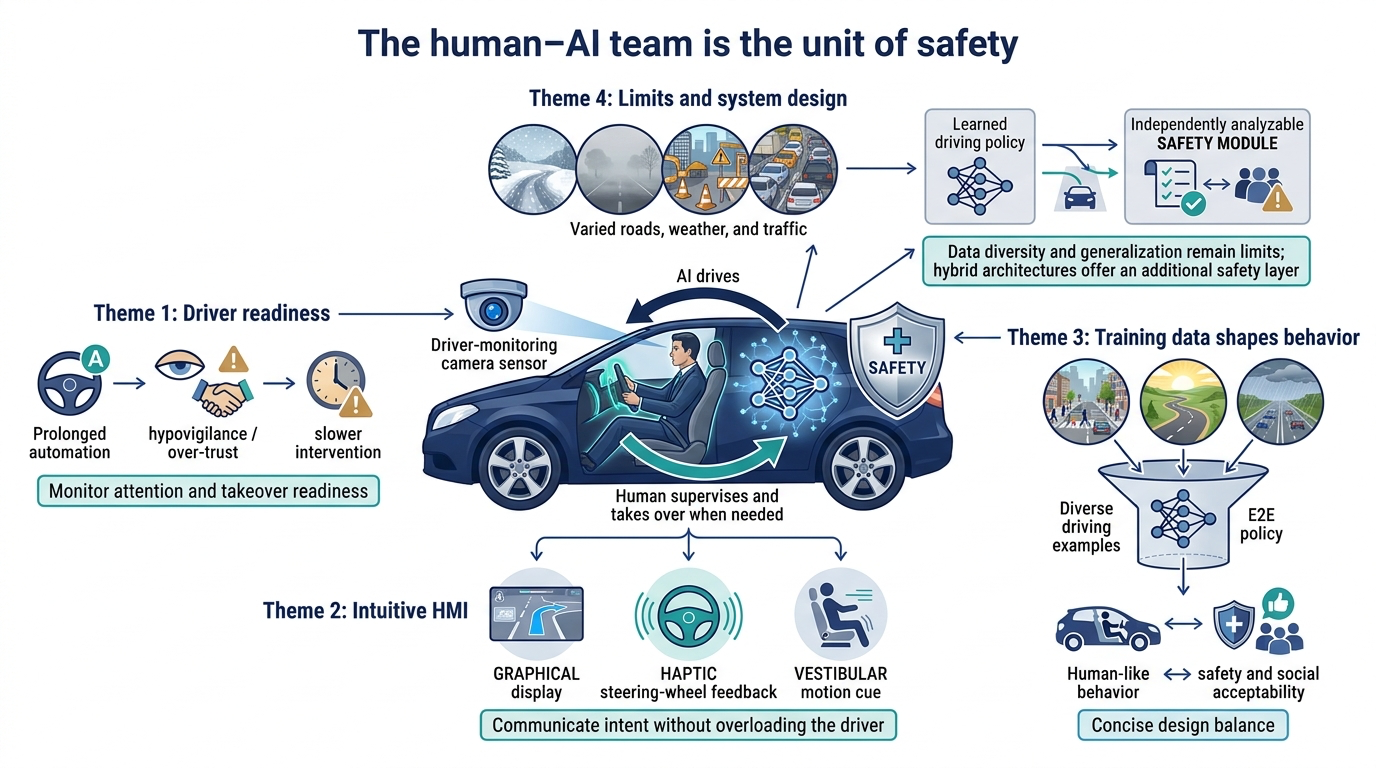}
    \caption{The human supervisor in supervised end-to-end driving: driver readiness, intent communication, the influence of training data on driving behaviour, and the limits that motivate hybrid architectures.}
    \label{fig:human}
\end{figure}

\subsection{Monitoring the Driver}
\label{sec:human:monitoring}

Because drivers will engage in non-driving tasks (and they will) the system must know what they are doing, so that risky behaviour is detected early and decisions in demanding situations can account for the driver's state. The technical approach is driver activity recognition with deep networks trained on in-cabin video. Most available datasets are recorded in simulators, without rain, poor weather, or the changes in illumination that break in-cabin vision; and the task vocabulary (phone use, drinking, reaching) understates what people actually do. 
To address these limitations, \citet{MoralesAlvarez2023} investigated the transferability of driver observation models from simulated to real driving environments, highlighting the challenges of generalising models across these conditions. They reported a substantial decrease in recognition accuracy from 85.7\% in simulation to 46.6\% in real-world conditions, which shows the limits of simulation-based training and the need for representative real-world data. Building on these findings, \citet{Sabry2024} extended the data collection by developing a driver monitoring dataset using a real-world automated driving research vehicle. The dataset captures driver activities under varying vehicle speeds and illumination conditions, providing a more representative basis for training and evaluating robust driver activity recognition systems. 
People riding with high levels of automation engage in behaviour they would not exhibit in front of their own television. Eye-tracking studies confirm the depth of immersion: given a phone task, participants stop looking at the road almost entirely \citep{MoralesAlvarez2020}; asked to find a small chocolate in a bag, one put his head inside the bag while the vehicle was moving. Gaze direction, dwell time on phone versus road, and glance transitions are all measurable with tracking glasses, and are the parameters a monitoring system needs to time a warning. Several studies estimate a person's psychological state (attention, stress, or trust) based on physiological indicators such as heart rate variability, electrodermal activity, or electroencephalography \citep{Deng2024AAP,Chen2025Systems,pena2025beyond, Li2025}; these signals have shown promising results and could eventually complement vision-based monitoring systems, but are very difficult to implement outside laboratory settings. Driver monitoring should therefore extend beyond detecting gaze direction or identifying secondary activities. A driver looking at the road may still lack sufficient situational awareness or cognitive readiness to intervene safely. This motivates a shift from monitoring observable driver states towards estimating the driver's actual capability to resume control. Such capability-oriented monitoring could support adaptive warnings and intervention strategies that account for both the driving situation and the driver's estimated readiness. This approach is consistent with the human-centred HMI design principles discussed by  \citet{OlaverriMonreal2017}, which emphasise the importance of considering human factors and contextual information when designing communication and assistance systems for automated driving. However, reliably estimating intervention capability in real-world conditions remains an open research challenge.

An eye-on-road request alone is probably insufficient; combining it with a hands-on request is more reliable; and any request must leave enough time to respond. Monitoring for drowsiness can trigger earlier and more pronounced warnings, though some drivers will fall asleep as well. Current production practice illustrates the problem: the system described in Section~\ref{sec:background} no longer requires hands on the wheel but does require the driver's eyes on the road, enforced by an in-cabin camera. A commercially sold device consists of a printed face with eyes that is placed in the camera's field of view and satisfies the attention check \citep{dacosta2026tesla}; weights that simulate hands on the wheel have existed for years. The reading of these devices is the important one: people go to this trouble because they are tired of being asked to supervise a system that, in their experience, does not need supervising. The existence of a market for circumvention devices provides evidence that some users are unwilling to comply with the intended supervisory role and may reflect over-trust in the automation.

The technology has moved from occasional assistance to sustained automated operation over much of a journey. This apparent reliability is what makes supervision difficult: failures become rare enough that drivers may cease to expect them. The knowledge gap is how to convince a user that a system which has never failed for them could fail, and in which situations it is likely to---which is why ``in which situations is the system likely to fail'' is, in our view, a question the field needs to answer, for engineers and for drivers.

\subsection{Interfaces That Inform Without Intruding}
\label{sec:human:hmi}

Drivers need to understand what the vehicle \emph{plans} to do and what it perceives. The person who buys such a vehicle does so in order not to be part of the driving loop, and does not want to be addressed by it continuously; natural-language narration of the kind (technically feasible) is unlikely to be welcome for this reason. The design target is an interface that is non-obstructive, immediately understandable, continuous in its feedback, and does not diminish the pleasure of using the car.

Field interviews show what such an interface achieves. In interviews conducted while driving with an end-to-end supervised system, users repeatedly explained their comfort in the same words: \emph{I know what the car is going to do}. The instrument display draws the planned path, shows the indicator, and, at a roundabout, renders the intended trajectory before the vehicle moves. A recorded entry into a fast road from a side street illustrates the layering. As a gap approaches, the display draws the path; the steering wheel begins to turn with no acceleration; then a small, perceptible acceleration follows; only after the driver has had time to corroborate the gap does the vehicle commit.
Intent is thus communicated over three channels (graphical, haptic through the wheel, and vestibular through the seat) in a sequence the driver can veto at any moment. Without that sequence the same manoeuvre would be uncomfortable and in some cases terrifying.

The importance of providing concise, context-relevant information extends beyond takeover warnings to the continuous supervision of automated driving. \citet{Hager2026}, investigated an in-vehicle HMI for conditionally automated platooning that communicated essential information about the automation state and inter-vehicle distance. In a driving-simulator study, the interface significantly reduced unnecessary manual interventions during normal platooning operation, with intervention rates approximately 80\% higher in the absence of the HMI. These results suggest that unobtrusive, task-relevant interfaces can support driver awareness and appropriate supervisory behaviour. 

The effectiveness of takeover assistance should not be assessed solely through reaction time. A rapid response does not necessarily imply that the driver has regained sufficient situational awareness to control the vehicle safely. Measures such as steering stability, trajectory deviation, and collision avoidance are therefore important complements to takeover time. Research on haptic guidance following takeover requests provides an additional perspective by examining how assistance during the transition can support the driver's control performance \citep{MoralesAlvarez2022}. Consequently, interface design should address not only when and how to alert the driver, but also how to support safe vehicle control after the intervention begins.

Although drivers in highly automated vehicles generally expect the system to manage most decisions reliably and autonomously, there are moments when they wish to actively guide the vehicle's behaviour. HMIs must evolve beyond displaying system status to actively accommodating user preferences. Just as a passenger might ask a taxi driver to take a slight detour, follow a specific vehicle, or choose a preferred route over the default, users should be able to communicate their intentions to an automated system naturally. Drivers do not need to be continuously involved in the operational driving loop to request a lane change, modify a driving style, or express a tactical preference. HMIs should facilitate intuitive, conversational interaction between the human and the vehicle, a paradigm where lightweight Large Language Models (LLMs) may provide one mechanism for such interaction \citep{vasquez2026eli}.

\subsection{Whose Driving Should the Car Imitate?}
\label{sec:human:style}

Production vehicles already offer selectable driving styles (eco, comfort, sport) and whether vulnerable road users should always be given priority is a live design decision. 
End-to-end training makes the choice of training population particularly important because driving styles differ substantially between drivers. A policy trained predominantly on aggressive or cautious driving may reproduce those tendencies, while strict rule compliance may conflict with locally negotiated traffic behaviour. The provenance and distribution of training behaviour become design variables. A further distinction is necessary between reproducing observed human driving behaviour and learning behaviour that is safe and socially acceptable. Naturalistic driving data capture informal traffic conventions, negotiation strategies, and interactions between road users \citep{Certad2024}, but may also contain unsafe or non-compliant behaviour \citep{Certad2023}. Studies of naturalistic yielding behaviour and vehicle interactions provide empirical information about how drivers negotiate traffic situations \citep{Gankhuyag2023}. However, behavioural representativeness alone does not establish which actions an automated system should reproduce. End-to-end training must therefore consider not only the diversity of driving behaviours represented in the data, but also how safety constraints and traffic regulations should govern the selection of acceptable actions. This matters most when transferring driving policies between countries or traffic environments with different interaction patterns.

\subsection{Generalisation and the Limits of Data}
\label{sec:human:generalisation}

Generalisation is constrained by the diversity and coverage of the training data, particularly for rare events and multi-agent interactions. Simulation can extend this coverage, although the sim-to-real gap is unresolved. Open software and reproducible evaluation pipelines may also improve the identification and comparison of failure modes across research groups. These considerations motivate hybrid architectures that combine the scalability of learned policies with components that can be independently analysed and validated.

\section{Open Science Requirements: Matching Architectures to Data, Tools and Benchmarks}
\label{sec:openscience}

Each architectural choice discussed in this paper moves the burden of evidence onto data. A modular stack is validated with labelled perception data and scenario tests; an end-to-end policy with long, synchronised driving logs and closed-loop simulation; a supervised system with data on the human as well as the machine. Because most of this evidence for supervised end-to-end driving is held inside manufacturers' fleets, safety claims can rarely be independently verified or reproduced. Open data, open-source tools and reproducible verification of models and methods are a precondition for the assurance arguments in Sections~\ref{sec:validation}--\ref{sec:human}. Table~\ref{tab:datamap} maps each architecture or component to the data it requires, representative open resources, and the main gaps.

\begin{table*}[pos=htbp]
\caption{Architectures and components, the data they require, representative open resources, and data gaps.}
\label{tab:datamap}
\footnotesize
\begin{tabularx}{0.98\textwidth}{>{\raggedright\arraybackslash}p{2.7cm}LLL}
\toprule
\textbf{Architecture / component} & \textbf{Data required} & \textbf{Open data, tools and benchmarks} & \textbf{Main gaps} \\
\midrule
Modular pipeline (Sec.~\ref{sec:background}) & Multi-sensor logs (camera, lidar, radar, GNSS/IMU); 3D boxes and tracks; HD maps; ego pose & KITTI \citep{geiger2012kitti}; nuScenes \citep{caesar2020nuscenes}; Waymo Open Dataset \citep{sun2020waymo}; Argoverse~2 \citep{wilson2021argoverse2}; Autoware \citep{kato2018autoware}, Apollo & Short clips from few cities; rare events and adverse weather; regional diversity \\
\addlinespace
End-to-end policy, imitation (Sec.~\ref{sec:background:training}) & Long synchronised multi-camera video with route, ego kinematics and control signals, at fleet scale & nuPlan \citep{caesar2021nuplan}; NAVSIM \citep{dauner2024navsim}; OpenDV-YouTube \citep{yang2024genad}; comma2k19 \citep{schafer2018comma2k19}; openpilot \citep{openpilot} & No open fleet-scale data; control signals rarely released; open-loop metrics weakly predict closed-loop behaviour \citep{li2024egostatus,zhai2023rethinking} \\
\addlinespace
End-to-end policy, long-tail refinement (Sec.~\ref{sec:background:training}) & Intervention, disengagement and near-miss episodes; reactive closed-loop simulation; scenario descriptions & CARLA \citep{dosovitskiy2017carla}; Bench2Drive \citep{jia2024bench2drive}; Waymax \citep{gulino2023waymax}; ScenarioNet \citep{li2023scenarionet}; ASAM OpenSCENARIO; public crash and disengagement reports & No open intervention logs from production systems; sim-to-real gap; simulator nondeterminism and lack of certification \\
\addlinespace
Prediction and causal modules (Sec.~\ref{sec:prediction}) & Long trajectories of interacting agents with context; pedestrian cues; intent labels & highD \citep{krajewski2018highd}, inD, rounD, exiD; INTERACTION \citep{zhan2019interaction}; Waymo Open Motion \citep{ettinger2021womd}; PIE \citep{rasouli2019pie}, JAAD; DoWhy \citep{sharma2020dowhy} & Intent and causal ground truth; cross-region transfer; data rich in vulnerable road users \\
\addlinespace
Explanations and language rationales (Sec.~\ref{sec:discussion:explain}) & Driving video paired with actions and rationales; perturbation tests & BDD-X \citep{kim2018bddx}; DriveLM \citep{sima2024drivelm}; DRAMA \citep{malla2023drama}; DriveBench \citep{xie2025vlmsready} & No ground truth for faithfulness to the policy; benchmarks answerable without the image \\
\addlinespace
Driver monitoring and HMI (Sec.~\ref{sec:human}) & In-cabin video, gaze and physiological signals synchronised with automation state, takeovers and outcomes & Drive\&Act \citep{martin2019driveact}; DMD \citep{ortega2020dmd}; \citep{Sabry2024}; SHRP2 NDS (restricted access) & On-road data with automation engaged; engagement over weeks; links to outcomes; privacy-preserving release \\
\addlinespace
In-service ADAS evaluation (Sec.~\ref{sec:evaluation}) & Lateral position, speed, acceleration and alert timing in naturalistic driving, across vehicles & Published test protocols \citep{EuroNCAP2026Safety,iso11270}; open foundation models \citep{sam3} & No open naturalistic ADAS performance data; no agreed lane-keeping quality metric \\
\addlinespace
Behavioural assurance (Sec.~\ref{sec:gooddriving:dcd}) & Compact records of state, perceived objects, applicable rules, predictions and intent, with software version & DCD \citep{bsi2021dcd}; ASAM OpenDRIVE, OpenSCENARIO, OpenLABEL as building blocks & No reference corpus; format not adopted; thresholds for careful and competent driving unknown \\
\addlinespace
Infrastructure and locality (Sec.~\ref{sec:validation:infra}) & Authoritative geo-referenced speed limits, signs and local rules & OpenStreetMap; road-authority data \citep{NSWGovernment2024SpeedAdviser}; GTSRB \citep{stallkamp2012gtsrb}; Mapillary Traffic Sign Dataset \citep{ertler2020mapillary} & Coverage, currency and local exceptions; no link between data quality and validation credit \\
\bottomrule
\end{tabularx}
\end{table*}

\subsection{Open Data: What Exists and What Is Missing}
\label{sec:openscience:data}
Open driving data are plentiful for perception and motion forecasting but thin for the questions this paper raises. Most public datasets consist of short clips from a small number of cities, with few rare events, little adverse weather and limited regional diversity, which matters given the locality arguments of Section~\ref{sec:validation:isa}. For end-to-end planning, open-loop evaluation on such datasets has proved unreliable: a planner given only the ego vehicle's own state can score competitively \citep{li2024egostatus,zhai2023rethinking}, which has motivated pseudo-closed-loop \citep{dauner2024navsim} and closed-loop \citep{jia2024bench2drive} benchmarks. Five categories of data are largely absent from the public domain: intervention and disengagement logs from supervised production systems; driver-state data synchronised with automation state and linked to outcomes; naturalistic performance data for everyday ADAS functions; ground truth for intent and for the faithfulness of explanations; and standardised behavioural records of the kind proposed by Digital Commentary Driving. Public crash reporting exists in some jurisdictions but is often redacted, which limits independent analysis. Independent initiatives such as White Box Autonomy \citep{zheng2026whitebox} begin to fill this gap by documenting and verifying real-world automated-driving events.

\subsection{Open-Source Tools and Reference Implementations}
\label{sec:openscience:tools}
Open-source stacks and simulators make it possible to reproduce methods and compare them on equal terms: modular stacks such as Autoware \citep{kato2018autoware} and Apollo, closed-loop simulators such as CARLA \citep{dosovitskiy2017carla} and Waymax \citep{gulino2023waymax}, scenario formats such as ASAM OpenSCENARIO and OpenDRIVE, and evaluation platforms such as OpenPAV \citep{zhou2026openpav}. Of particular relevance here is openpilot \citep{openpilot}, an open-source Level~2 driver-assistance system with a learned driving model running on production vehicles, which allows supervised learned driving to be studied without depending on a manufacturer. The methods illustrated in this paper were chosen with reproducibility in mind: the portable lane-support measurement relies on an open foundation model \citep{sam3} and no proprietary CAN access, and the causal lane-change analysis uses a public dataset \citep{krajewski2018highd} and an open causal-inference library \citep{sharma2020dowhy}. Simulation-based validation will only be accepted if the simulators themselves are open to scrutiny, versioned and certified (Section~\ref{sec:validation:trends}).

\subsection{Verification and Reproducibility of Models and Methods}
\label{sec:openscience:repro}
Reproducible safety claims require reporting practices as much as data. We propose a minimum set for research on learned driving and supervised automation: (i) release code, configurations and, where possible, model weights; where the model is proprietary, release the evaluation harness and scenario files in an open format; (ii) report closed-loop results alongside open-loop metrics; (iii) report simulator and software versions and random seeds, and repeat stochastic evaluations with reported variance, given the nondeterminism noted in Section~\ref{sec:validation:trends}; (iv) document the provenance and distribution of training behaviour with datasheets, since whose driving a policy imitates is a design variable (Section~\ref{sec:human:style}); (v) report faithfulness tests for any explanation offered (Table~\ref{tab:explanations}); and (vi) tag in-service evidence with the software version that produced it, so that results survive over-the-air updates. These practices would allow the research community, and regulators, to reproduce rather than accept the safety claims on which deployment decisions rest.

\section{Discussion: Open Questions and a Research Agenda}
\label{sec:discussion}

We organise the discussion around research gaps. For each we state what is currently established, what remains open, and what research would close the gap. Table~\ref{tab:agenda} summarises the conclusions.
 
\subsection{What Is Good Driving, and Against Whom Is Safety Measured?}
\label{sec:discussion:baseline}
 
Developers of supervised end-to-end systems report that their vehicles are several times safer than the average driver, measured in kilometres per crash. We argue that this comparison is not the right one. The average driver includes the drunk, the distracted and the untrained; most drivers are considerably better than the average, and the relevant concern for someone deciding whether to let the system drive is whether it is better than \emph{they} are. This requires a definition of good driving \citep{waymo2026sober} that is operational rather than statistical, the careful and competent driver of Section~\ref{sec:gooddriving}: what such a driver would be expected to do in a given situation, compared with what the vehicle reports it perceived, predicted and intended, so that behaviour can be assessed from standardised, compact in-vehicle data such as Digital Commentary Driving \citep{reed2024careful, bsi2021dcd}. Minor collisions need an agreed weighting; a system that eliminates severe crashes but increases low-speed contacts is a different proposition from one that does neither, and current crash-rate statistics do not distinguish them. And the metric must be conditioned on situation, because ``when is it better than the human'' is not a number but a map of situations, which is also the map of where the human is still needed.
 
\emph{Research gap:} a shared, situation-conditioned definition of competent driving; benchmark scenario sets in which both human and system behaviour can be scored against it; and crash-severity-weighted metrics that can be computed from in-service data.
 
\subsection{Is the Supervisor Role Acceptable, and How Is It Monitored?}
\label{sec:discussion:supervisor}
 
%Here the workshop held a genuine tension, and it should be reported as one. The human-factors evidence in Section~\ref{sec:human} says that humans are poor supervisors of reliable automation, and that Level~2 and Level~3 are the dangerous region of the spectrum. The deployment evidence in Section~\ref{sec:background} says that hundreds of thousands of drivers are supervising an end-to-end system for most of their driving time, apparently with acceptable outcomes. Both are true, and the resolution is empirical: the field has been arguing from simulator studies and short on-road trials, and now has, for the first time, a large population of supervisors whose behaviour and outcomes are being logged continuously.

There is a tension between established human-factors findings and recent deployment experience. Laboratory and simulator studies show that sustained supervision of reliable automation is difficult, while supervised end-to-end systems are being used on a larger scale on public roads. Quantifying how drivers supervise these systems in routine use and how supervision affects safety outcomes requires field data; deployment scale alone is not evidence.
 
 The supervisor's engagement depends on knowing what the vehicle will do next; intent can be communicated peripherally, haptically and through motion without narration; eyes-on-road enforcement is being circumvented by a growing market of devices; and users need to be told in which situations the system is likely to fail, because their own experience will not tell them. It is unclear if the supervisory role is acceptable as the system approaches, but never reaches perfection, and if increasing automation reliability continues to improve safety as the human becomes responsible for increasingly rare interventions.
 
\emph{Research gap:} in-service driver-monitoring data linked to outcomes, made available for independent analysis; driver-activity datasets recorded on the road rather than in simulators; HMI designs evaluated on \emph{sustained} engagement over weeks rather than on takeover time in a single session; and a defensible answer to what drivers should be trained to expect.
 
\subsection{Explainability and Safety Guarantees for Learned Policies}
\label{sec:discussion:explain}
 
Section~\ref{sec:validation} needs to know what happened when an end-to-end system failed; Section~\ref{sec:prediction} shows how contextual and causal representations can provide interpretable predictions without requiring all relevant relationships to be implicitly learned from raw data; Section~\ref{sec:human} needs the vehicle's plan, not its perception, communicated to the driver. End-to-end developers already generate auxiliary outputs (segmentation, predicted paths, attention or saliency maps, language rationales) but none of these is known to be 100\% faithful to the policy's computation. An explanation can be plausible to a human without being faithful to the model~\citep{jacovi2020faithfully}, and a rationale that is plausible but unfaithful is worse than none because it manufactures trust.

Sources of explanation differ in how directly they reflect the policy's decision (Table~\ref{tab:explanations}). The planned path on the display shows what the vehicle intends to do but not why, and the outputs meant to explain why are those whose faithfulness is least certain. Masking showed that more than half of the regions a driving network attended to did not actually influence its output~\citep{kim2017causalattention}; some saliency methods are independent of the model they are meant to explain~\citep{adebayo2018sanity}; and vision-language models asked about driving scenes often answer from general knowledge or textual cues rather than from the images, many scoring almost as well with no image at all~\citep{xie2025vlmsready}. Where an explanation points to part of the input, its faithfulness can be tested by removing that part and measuring how much the output changes~\citep{petsiuk2018rise}.
Digital Commentary Driving (Section~\ref{sec:gooddriving:dcd}) relies on what the vehicle reports it perceived, predicted and intended, therefore its value for assurance also depends on whether those reports are faithful.

\begin{table}[pos=htbp]
\caption{Sources of explanation in an end-to-end driving system.\label{tab:explanations}}
\small

\begin{tabularx}{\textwidth}{>{\raggedright\arraybackslash}p{4.2cm}LL}
\toprule
\textbf{Source} & \textbf{Link to decision} & \textbf{Faithfulness test} \\
\midrule
Planned path on the display (Section~\ref{sec:human:hmi}) & Shows what the vehicle intends to do, not why & Compare the drawn path with the executed one \\
Auxiliary outputs: segmentation, motion forecasting (Section~\ref{sec:background:consequences}) & Produced alongside the policy; may differ from what it used & Indirectly, by checking that behaviour is consistent with them \\
Attention or saliency maps (Section~\ref{sec:background:consequences}) & Derived from the policy, but not proof of what it used & Remove the highlighted regions and measure the change in the trajectory \\
Language rationales (Sections~\ref{sec:background:consequences}, \ref{sec:gooddriving:dcd}) & A generated summary; may be plausible without being faithful & Compare the stated reasoning with the action \\
Separate prediction modules (Section~\ref{sec:prediction}) & Explain their own computation, not the policy's & Check agreement with the policy's behaviour \\
\bottomrule
\end{tabularx}
\end{table}

We favour hybrid architectures in which a learned driving policy is complemented by modules for prediction, intent communication and ODD monitoring. These modules can limit what the policy does, monitor its behaviour, or explain its decisions, providing another way to support safety.
 
\emph{Research gap:} faithfulness tests for all explanations generated by end-to-end systems (Table~\ref{tab:explanations}); causal analysis of learned policies, extending the structural-causal-model approach from prediction to decision making; formal verification of the modules and runtime monitoring of the learned one; and statistical assurance methods for rare events.
 
\subsection{Open Data, Tools and Reproducibility}
\label{sec:discussion:data}

Section~\ref{sec:openscience} showed that every learned component is bounded by its training data, that the data most needed for assurance are proprietary, and that modern vehicles can already record and upload them, so data sharing is a policy decision rather than a technical limitation. Safety analysis should therefore not depend solely on the data manufacturers choose to publish.
 
\emph{Research gap:} a regulatory framework for access to fleet driving logs, with privacy-preserving analysis; certified simulation tools and validated scenario generators; public datasets with the rare events, multi-agent interactions, and weather that current datasets lack; and reference implementations that allow independent reproduction of safety claims.
 
\subsection{Infrastructure, Locality and Social Norms}
\label{sec:discussion:infra}
 
The intelligent-speed-assistance case of Section~\ref{sec:validation:isa} and the discussion of the social-norms are the same problem on two scales. On a small scale, a vehicle being asked to infer from a camera what the road authority already knows: the posted limit. On a large scale, a vehicle trained in one city being asked to drive in another whose effective rules are social. In the two cases, we ask whether the road should adapt to the vehicle instead of only the reverse.
 
A geo-referenced, authoritative database of speed limits maintained by the road authority would remove an entire category of validation effort; harmonised signage across jurisdictions would remove another; but old cities, local exceptions and the sheer number of small authorities make both slow and expensive. Social norms cannot be published at all; they can only be learned or reasoned about, which is why neuro-symbolic approaches that combine traffic law, local custom and common sense are on the agenda.
 
\emph{Research gap:} a specification of which ODD elements should be provided as authoritative infrastructure data and how vehicle validation gives credit for them; cost, benefit analysis of infrastructure modernisation against per-vehicle validation; and methods for representing and learning local norms, with transfer evaluation across cities.
 
\subsection{Insurance, Liability and Regulation}
\label{sec:discussion:regulation}
 
The technology is deployed under a Level~2 classification that assigns full responsibility to a driver who is, for most of the trip, not driving. The definitions separating L2, ``L2++'' and L3 are not agreed; the validation evidence that authorities should require is not agreed; whether the function is a safety feature or a convenience feature is not agreed; and the beneficiaries who would gain most, such as older drivers whose own capabilities are declining, are not those for whom the supervisory role is designed. %Regulation is, as industry practitioners note, being written while the systems it governs are already on the road.
 
\emph{Research gap:} evidence-based regulatory categories that describe the deployed products; liability frameworks for shared control that reflect measured, not nominal, allocation of the driving task; and evaluation of the benefit to specific populations.
 
\begin{table}[pos=htbp]
  \caption{Research agenda: current understanding,
  and what remains to be done.}
  \label{tab:agenda}
  \centering
  \small
  \begin{tabular}{@{}p{2.5cm}p{6.0cm}p{6.0cm}@{}}
    \toprule
    \textbf{Question} & \textbf{Current understanding} & \textbf{Open / Research gap} \\
    \midrule
    Good driving and the safety baseline & Average-driver comparison is inadequate; competent-driver notion is operational; minor crashes need weighting & Situation-conditioned definition; scenario benchmarks scoring human and system alike; severity-weighted in-service metrics \\
    \addlinespace
    Acceptability and monitoring of the supervisor & Engagement depends on knowing intent; peripheral/haptic/motion cues work; enforcement is circumvented; users must be told where the system fails & In-service monitoring linked to outcomes; on-road activity datasets; HMI evaluated on sustained engagement; whether near-perfect reliability is more or less safe \\
    \addlinespace
    Explainability and assurance & Hybrid learned-policy + explicit-module architectures favoured; prediction can support interpretability & Faithfulness of E2E explanations; causal analysis of policies; formal verification of modules, runtime monitoring of policy; rare-event statistics \\
    \addlinespace
    Open data, tools, reproducibility &
    Fleet data is the discovery mechanism but is proprietary; simulation needed and unreliable; open ecosystems proposed & Regulated access to fleet logs; certified simulators and scenario generators; public rare-event datasets; reproducible safety claims; reporting standards \\
    \addlinespace
    Infrastructure, locality, norms & ISA failures are largely infrastructure failures; authoritative data would cut validation cost; local norms cannot be published & ODD elements as infrastructure data; cost--benefit of modernisation; learning and transferring local norms and driving cultures \\
    \addlinespace
    Insurance, liability, regulation & L2 classification does not describe deployed use; definitions and evidence requirements not agreed; safety-vs-convenience unresolved & Regulatory categories for deployed products; liability for shared control; benefit to specific populations \\
    \bottomrule
  \end{tabular}
\end{table}
 
\subsection{Scope and Limitations}
\label{sec:discussion:method}
This is a perspective paper, not a systematic review. Its arguments draw on the authors' research and industrial experience across several countries and on the discussions acknowledged in the Introduction, and its claims should be weighed accordingly. The overview of data, tools and benchmarks in Section~\ref{sec:openscience} is representative, not exhaustive. 
The causal lane-change analysis summarises published work, whereas the in-service lane-support comparison is preliminary. Both are included to illustrate methods and do not establish general conclusions.

% ===============================================================
\section{Conclusion}
\label{sec:conclusion}
 
Driver assistance is undergoing its largest change since it began. Supervised end-to-end driving systems perform essentially the whole driving task on public roads in ordinary production cars, in both left- and right-hand traffic, for the large majority of trip time, with a human who is legally the driver and functionally a supervisor. This paper has examined what this change demands of the way such systems are built, validated, supervised and evaluated, and what it demands of the openness of the evidence on which those judgements rest.
 
The move from modular pipelines to learned policies shifts validation from deterministic engineering to probabilistic assurance, and the standards, metrics and tools of validation are being rebuilt to match while the systems are already deployed. Prediction, monitoring and explanation mechanisms can complement an end-to-end policy by providing independently inspectable information for validation and communication. Their value ultimately depends on whether their outputs are faithful to the underlying policy and demonstrably improve assurance. The safety of a supervised system is the safety of the human--AI pair, and the human-factors evidence that people supervise reliable automation poorly must be reconciled with the deployment evidence that they are doing so at scale; the reconciliation is empirical and requires access to the data that deployed fleets already collect. The everyday assistance functions on which most of the fleet still relies can and should be measured objectively, in naturalistic driving, without infrastructure or manufacturer cooperation, and preliminary measurements suggest  differences between implementations.

A common thread runs through these arguments: the evidence needed to judge supervised end-to-end driving is data, and most of it is closed. Mapping each architecture to its data requirements shows that open resources are plentiful for perception and motion forecasting but scarce for interventions, driver state linked to outcomes, naturalistic ADAS performance and explanation faithfulness. The contributions most likely to move the field forward are therefore open ones: on-road datasets recorded with supervised systems engaged, closed-loop benchmarks that predict real-world behaviour, open and certified simulators, standardised behavioural records, and reporting practices that make every safety claim reproducible.

We have argued that hybrid architectures are the most promising direction: learned driving policies complemented by components for prediction, intent communication, ODD monitoring and in-service evaluation. The research agenda in Section~\ref{sec:discussion} identifies the evidence needed to determine when and how such combinations improve assurance, explainability and safety in public-road deployment.
 
\section*{Data Availability}
The lane-support data shown in Fig.~\ref{fig:lateral} are part of ongoing work that will be published separately.

\section*{Declaration of competing interest}
The authors declare that they have no known competing financial interests or personal relationships that could have appeared to influence the work reported in this paper.

\section*{Funding}
This work was partially supported by the Australian Research Council (ARC) Industrial Transformation Training Centre for Automated Vehicles in Rural and Remote Regions (AVR3), grant number IC230100001.

\section*{CRediT authorship contribution statement}
\textbf{Julie Stephany Berrio-Perez:} Conceptualization, Writing -- original draft, Writing -- review \& editing, Visualization, Project administration.
\textbf{Eduardo Nebot:} Conceptualization, Writing -- original draft, Writing -- review \& editing.
\textbf{Javier Ibanez-Guzman:} Conceptualization, Writing -- original draft, Writing -- review \& editing.
\textbf{Nick Reed:} Conceptualization, Writing -- original draft, Writing -- review \& editing.
\textbf{Miguel Angel Sotelo:} Conceptualization, Writing -- original draft, Writing -- review \& editing.
\textbf{Cristina Olaverri-Monreal:} Conceptualization, Writing -- original draft, Writing -- review \& editing.
\textbf{Zuduo Zheng:} Writing -- review \& editing.
\textbf{Prathyush P Menon:} Writing -- review \& editing.
\textbf{Walter Zimmer:} Writing -- review \& editing.
\textbf{Juan Medina-Lee:} Writing -- review \& editing.
\textbf{Mona Mirzaie:} Writing -- review \& editing.
\textbf{Tyson Phillips:} Writing -- review \& editing.
\textbf{Mao Shan:} Conceptualization, Writing -- original draft, Writing -- review \& editing.

\section*{Declaration of generative AI and AI-assisted technologies in the writing process}
During the preparation of this work, the authors used Claude (Anthropic) and Gemini (Google) to improve wording and grammar. After using these tools, the authors reviewed and edited the content as needed and take full responsibility for the content of the published article.

\printcredits

\bibliographystyle{cas-model2-names}

\bibliography{references}


%% Author biographies (CAS format).
%% To add a photo, put the image path inside \bio{...}, e.g. \bio{figs/berrio}.

\bio{}
\textbf{Julie Stephany Berrio Perez} is a Senior Lecturer in the School of Mechanical and Mining Engineering at The University of Queensland, Australia. She received her PhD from the Australian Centre for Field Robotics at the University of Sydney, with a thesis on scene understanding and map maintenance for autonomous vehicles. Her research covers multimodal perception and sensor fusion (camera, LiDAR and radar) for scene understanding, semantic occupancy prediction, HD mapping, domain adaptation and anomaly detection, as well as long-term mapping and localisation. She contributed to on-road connected and automated vehicle trials in complex urban settings supported by Transport for NSW, and co-created the STU, Mixed Signals and Panoptic-CUDAL datasets for road anomaly segmentation, V2X cooperative perception and driving in adverse rural conditions.
\endbio

\bio{}
\textbf{Eduardo Nebot} is an Emeritus Professor at The University of Sydney, Australia. He co-founded the Australian Centre for Field Robotics (ACFR), served as its Managing Director from 2010 to 2020, and established its Intelligent Transport Systems group in 2010. His research focuses on navigation and automation for robotics, including high-integrity navigation, localisation and mapping, autonomous safety systems, and automation for the mining and automotive industries. He led the automation programme of the Cooperative Research Centre for Mining Technology and Equipment, co-founded Acumine Pty Ltd, and introduced mining safety technology deployed in mines worldwide. He is a Fellow of the IEEE and of the Australian Academy of Technological Sciences and Engineering.
\endbio

\bio{}
\textbf{Javier Ibanez Guzman} is Corporate Expert on Autonomous Systems at Renault Group, France, and Co-Director of SIVALab, a joint laboratory between CNRS, UTC Compi\`egne and Renault. He received an MSEE from the University of Pennsylvania as a Fulbright Scholar and a PhD from the University of Reading, UK, under an SERC Fellowship. He was previously a Senior Scientist at SimTech (A*STAR), Singapore, where he led autonomous ground vehicle initiatives, and a Visiting Scholar at UC Berkeley working on connected vehicles. He represents France in ISO groups on automated vehicles and artificial intelligence, serves as an expert for EU research programmes, and is a Chartered Engineer, a Fellow of the IET and an IEEE Life Member.
\endbio

\bio{}
\textbf{Nick Reed} is the founder of Reed Mobility, an independent research consultancy on the future of mobility, and a Visiting Professor at the University of Surrey, UK. He holds an MA from the University of Cambridge and a DPhil from the University of Oxford, and is a Chartered Scientist and psychologist. He spent 13 years leading human factors and driver behaviour research at TRL, where he became Academy Director, before serving as Head of Mobility R\&D at Bosch. In 2021 he was appointed the first Chief Road Safety Adviser to National Highways, and he has served as an international expert on the European Commission panel on the ethics of connected and automated vehicles.
\endbio

\bio{}
\textbf{Miguel \'Angel Sotelo} is a Full Professor in the Department of Computer Engineering at the University of Alcal\'a, Spain. He received his PhD in Electrical Engineering from the University of Alcal\'a in 2001 and an MBA from the European Business School in 2008. His research interests include driverless vehicles, cooperative intelligent transportation systems and traffic technologies, and he has authored more than 300 refereed papers. He was President of the IEEE Intelligent Transportation Systems Society (2018--2019) and Editor-in-Chief of the IEEE Intelligent Transportation Systems Magazine (2014--2017). He is an IEEE Fellow and received the 2022 IEEE ITSS Outstanding Research Award.
\endbio

\bio{}
\textbf{Cristina Olaverri Monreal} is a Full Professor at Johannes Kepler University Linz (JKU), Austria, where she heads the Department of Intelligent Transportation Systems. She studied computational linguistics, computer science, and phonetics at Ludwig Maximilian University of Munich and earned her doctorate in 2006 in collaboration with BMW. Before joining JKU, she held positions in industry and academia across several European countries and the United States. Her research focuses on improving the efficiency and effectiveness of transportation systems, including automated driving, human factors, and the application of wireless communication and sensor technologies to intelligent mobility. She served as President of the IEEE Intelligent Transportation Systems Society (2022–2023) and currently holds the position of IEEE Division IX Director (2026–2027), serving on the IEEE Board of Directors. Her contributions have been recognized with the 
2017 IEEE Meritorious Achievement in Continuing Education and 
2023 IEEE MGA Diversity \& Inclusion Awards.
\endbio

\bio{}
\textbf{Zuduo Zheng} is a Professor in the School of Civil Engineering at The University of Queensland, Australia, and an Australian Research Council (ARC) Future Fellow. He received his PhD from Arizona State University and is a former ARC DECRA Fellow. His research interests include traffic flow theory and operations with emerging technologies, travel behaviour, transport planning and modelling, traffic safety, and advanced data analysis in transport engineering. He has served as an editor or editorial board member for journals including Transportation Research Part B, Transportation Research Part C, Analytic Methods in Accident Research and IEEE Transactions on Intelligent Transportation Systems, and is a member of the ARC College of Experts.
\endbio

\bio{}
\textbf{Prathyush P. Menon} holds the Personal Chair in Control and Autonomy in the Department of Engineering at the University of Exeter, UK, where he is Deputy Head of Engineering and Deputy Director of the Centre for Future Clean Mobility. He leads the Cooperative Robotics and Autonomous Networks (CRANE) Lab. He received his PhD in Control Engineering from the University of Leicester in 2007, where he was later an ESA-funded Postdoctoral Research Associate and a Lecturer, before joining Exeter in 2010. His research covers autonomy across land, air and sea, decision-making under uncertainty, information-driven planning, cooperation and resilience in multi-agent systems, and the validation of autonomous systems.
\endbio

\bio{}
\textbf{Walter Zimmer} is a Postdoctoral Researcher at the University of California, Los Angeles (UCLA), USA. He received his PhD from the Technical University of Munich (TUM) in 2025 and his MSc in Computer Science from TUM, including a research stay at TU Delft. He previously worked as an Autonomous Systems Engineer at STTech GmbH and was a Visiting Scholar at the Laboratory for Intelligent and Safe Automobiles at UC San Diego. His research focuses on cooperative autonomous driving, 3D perception and 3D foundation models, and he led the creation of the A9 and TUMTraf roadside and V2X perception datasets. He has authored over 30 publications and received the IEEE ITSS Best Student Paper Award (2023) and a Best Dissertation Award (2025).
\endbio

\bio{}
\textbf{Juan Medina Lee} is an Assistant Professor in the Department of Computer Science and Engineering at the University of Puerto Rico--Mayag\"uez. He received his PhD in Automation and Robotics (\emph{cum laude}) from the Universidad Polit\'ecnica de Madrid, Spain, in 2023, an MSc in Engineering from Universidad del Valle, Colombia, and a degree in Electronic Engineering from the University of Quind\'io, Colombia. He was a researcher in the AUTOPIA group at the Centre for Automation and Robotics in Madrid, working on European and national autonomous-driving projects, and taught digital systems and robotics at the University of Quind\'io from 2012 to 2018. His research interests include decision-making and human factors in autonomous vehicles, embedded systems and mobile robotics.
\endbio

\bio{}
\textbf{Mona Mirzaie} is a doctoral researcher at the Institute for Information Processing (TNT), Leibniz University Hannover, Germany, which she joined in 2023. She received her MSc in Computer Science from Leibniz University Hannover with a thesis on industrial anomaly detection. From 2022 to 2023 she was a research assistant at the Technical University of Braunschweig, working with Volkswagen Group Research on sensor fusion for autonomous driving. Her doctoral research focuses on interpretable methods for end-to-end autonomous driving, and her interests include imitation and reinforcement learning, interpretable and trustworthy AI, visual perception and scene understanding, risk anticipation, and safety-critical, uncertainty-aware decision-making.
\endbio

\bio{}
\textbf{Tyson Phillips} is a Senior Lecturer and Director of Teaching and Learning in the School of Mechanical and Mining Engineering at The University of Queensland, Australia, and an affiliate of the School's Future Autonomous Systems and Technology group. He received his PhD from The University of Queensland. His research focuses on perception and automation for field machines, including LiDAR and point-cloud perception, object pose estimation, sensor registration and terrain mapping, with application to autonomous mining equipment such as excavators, rope shovels, haul trucks and bulldozers. He has worked on industry-funded projects with Caterpillar on articulated truck automation and electric rope shovel operator assistance, as well as projects with Fortescue Metals Group and the Australian Coal Association Research Program.
\endbio

\bio{}
\textbf{Mao Shan} is a Senior Research Fellow at the Australian Centre for Robotics, The University of Sydney, Australia. He received his BSc in Electrical Engineering from Shaanxi University of Science and Technology, China, and his MSc in Automation and Manufacturing Systems and PhD in Engineering from The University of Sydney in 2009 and 2014, respectively. He was a Research Fellow at Nanyang Technological University, Singapore, from 2016 to 2017. His research interests include localisation and tracking, vehicle-to-everything (V2X) communication, cooperative intelligent transport systems, cooperative perception and its safety evaluation, multi-sensor fusion, 3D semantic occupancy prediction and pedestrian trajectory prediction.
\endbio
% Author biography without author photo.
% Author biography. Author biography. Author biography.
% Author biography. Author biography. Author biography.
% Author biography. Author biography. Author biography.
% Author biography. Author biography. Author biography.
% Author biography. Author biography. Author biography.
% Author biography. Author biography. Author biography.
% Author biography. Author biography. Author biography.
% Author biography. Author biography. Author biography.
% Author biography. Author biography. Author biography.

% \bio{figs/cas-pic1}
% Author biography with author photo.
% Author biography. Author biography. Author biography.
% Author biography. Author biography. Author biography.
% Author biography. Author biography. Author biography.
% Author biography. Author biography. Author biography.
% Author biography. Author biography. Author biography.
% Author biography. Author biography. Author biography.
% Author biography. Author biography. Author biography.
% Author biography. Author biography. Author biography.
% Author biography. Author biography. Author biography.

% \vskip3pc

% \bio{figs/cas-pic1}
% Author biography with author photo.
% Author biography. Author biography. Author biography.
% Author biography. Author biography. Author biography.
% Author biography. Author biography. Author biography.
% Author biography. Author biography. Author biography.

\end{document}